\documentclass{article}

\usepackage[final,nonatbib]{nips_2016}

\usepackage{csquotes}
\usepackage[utf8]{inputenc} 
\usepackage[T1]{fontenc}    
\usepackage[hidelinks]{hyperref}  
\usepackage{url}            
\usepackage{booktabs}       
\usepackage{amsfonts}       
\usepackage{nicefrac}       
\usepackage{microtype}      
\usepackage{amsmath}
\usepackage{graphicx}
\usepackage{subcaption}
\usepackage{url}
\usepackage[labelfont=bf]{caption}

\usepackage{xcolor}
\usepackage{listings}

\title{Model Evaluation, Model Selection, and Algorithm Selection in Machine Learning}

\author{
  Sebastian Raschka\\
  University of Wisconsin--Madison\\
  Department of Statistics\\
  November 2018\\
  \texttt{sraschka@wisc.edu}
}

\begin{document}

\maketitle

\begin{abstract}

The correct use of model evaluation, model selection, and algorithm selection techniques is vital in academic machine learning research as well as in many industrial settings. This article reviews different techniques that can be used for each of these three subtasks and discusses the main advantages and disadvantages of each technique with references to theoretical and empirical studies. Further, recommendations are given to encourage best yet feasible practices in research and applications of machine learning. Common methods such as the holdout method for model evaluation and selection are covered, which are not recommended when working with small datasets. Different flavors of the bootstrap technique are introduced for estimating the uncertainty of performance estimates, as an alternative to confidence intervals via normal approximation if bootstrapping is computationally feasible. Common cross-validation techniques such as leave-one-out cross-validation and $k$-fold cross-validation are reviewed, the bias-variance trade-off for choosing $k$ is discussed, and practical tips for the optimal choice of $k$ are given based on empirical evidence. Different statistical tests for algorithm comparisons are presented, and strategies for dealing with multiple comparisons such as omnibus tests and multiple-comparison corrections are discussed. Finally, alternative methods for algorithm selection, such as the combined $F$-test 5x2 cross-validation and nested cross-validation, are recommended for comparing machine learning algorithms when datasets are small.

\end{abstract}

\newpage

\tableofcontents

\newpage

\section{Introduction: Essential Model Evaluation Terms and Techniques}
\label{sec:introduction}

Machine learning has become a central part of our life -- as consumers, customers, and hopefully as researchers and practitioners. Whether we are applying predictive modeling techniques to our research or business problems, I believe we have one thing in common: We want to make "good" predictions. Fitting a model to our training data is one thing, but how do we know that it generalizes well to unseen data? How do we know that it does not simply memorize the data we fed it and fails to make good predictions on future samples, samples that it has not seen before? And how do we select a good model in the first place? Maybe a different learning algorithm could be better-suited for the problem at hand?

Model evaluation is certainly not just the end point of our machine learning pipeline. Before we handle any data, we want to plan ahead and use techniques that are suited for our purposes. In this article, we will go over a selection of these techniques, and we will see how they fit into the bigger picture, a typical machine learning workflow.

\subsection{Performance Estimation: Generalization Performance vs. Model Selection}

Let us consider the obvious question, "How do we estimate the performance of a machine learning model?" A typical answer to this question might be as follows: "First, we feed the training data to our learning algorithm to learn a model. Second, we predict the labels of our test set. Third, we count the number of wrong predictions on the test dataset to compute the model's prediction accuracy." Depending on our goal, however, estimating the performance of a model is not that trivial, unfortunately. Maybe we should address the previous question from a different angle: "Why do we care about performance estimates at all?" Ideally, the estimated performance of a model tells how well it performs on unseen data -- making predictions on future data is often the main problem we want to solve in applications of machine learning or the development of new algorithms.  Typically, machine learning involves a lot of experimentation, though -- for example, the tuning of the internal knobs of a learning algorithm, the so-called hyperparameters. Running a learning algorithm over a training dataset with different hyperparameter settings will result in different models. Since we are typically interested in selecting the best-performing model from this set, we need to find a way to estimate their respective performances in order to rank them against each other.  

Going one step beyond mere algorithm fine-tuning, we are usually not only experimenting with the one single algorithm that we think would be the "best solution" under the given circumstances. More often than not, we want to compare different algorithms to each other, oftentimes in terms of predictive \textit{and} computational performance. Let us summarize the main points why we evaluate the predictive performance of a model:

\begin{enumerate}
\item We want to estimate the generalization performance, the predictive performance of our model on future (unseen) data.
\item We want to increase the predictive performance by tweaking the learning algorithm and selecting the best performing model from a given hypothesis space.
\item We want to identify the machine learning algorithm that is best-suited for the problem at hand; thus, we want to compare different algorithms, selecting the best-performing one as well as the best performing model from the algorithm's hypothesis space.
\end{enumerate}

Although these three sub-tasks listed above have all in common that we want to estimate the performance of a model, they all require different approaches. We will discuss some of the different methods for tackling these sub-tasks in this article.

Of course, we want to estimate the future performance of a model as accurately as possible. However, we shall note that biased performance estimates are perfectly okay in model selection and algorithm selection if the bias affects all models equally. If we rank different models or algorithms against each other in order to select the best-performing one, we only need to know their "relative" performance. For example, if all performance estimates are pessimistically biased, and we underestimate their performances by 10\%, it will not affect the ranking order. More concretely, if we obtaind three models with prediction accuracy estimates such as

M2: 75\% > M1: 70\% > M3: 65\%,

we would still rank them the same way if we added a 10\% pessimistic bias:

M2: 65\% > M1: 60\% > M3: 55\%.

However, note that if we reported the generalization (future prediction) accuracy of the best ranked model (M2) to be 65\%, this would obviously be quite inaccurate. Estimating the absolute performance of a model is probably one of the most challenging tasks in machine learning.

\subsection{Assumptions and Terminology}

Model evaluation is certainly a complex topic. To make sure that we do not diverge too much from the core message, let us make certain assumptions and go over some of the technical terms that we will use throughout this article.

\paragraph{i.i.d.} We assume that the training examples are i.i.d (independent and identically distributed), which means that all examples have been drawn from the same probability distribution and are statistically independent from each other. A scenario where training examples are not independent would be working with temporal data or time-series data.

\paragraph{Supervised learning and classification.} This article focusses on supervised learning, a subcategory of machine learning where the target values are known in a given dataset. Although many concepts also apply to regression analysis, we will focus on classification, the assignment of categorical target labels to the training and test examples.

\paragraph{0-1 loss and prediction accuracy.} In the following article, we will focus on the prediction accuracy, which is defined as the number of all correct predictions divided by the number of examples in the dataset. We compute the prediction accuracy as the number of correct predictions divided by the number of examples $n$. Or in more formal terms, we define the prediction accuracy ACC as

\begin{equation}
\text{ACC} = 1 - \text{ERR},
\end{equation}

where the prediction error, ERR, is computed as the expected value of the 0-1 loss over $n$ examples in a dataset $S$:

\begin{equation}
\text{ERR}_S = \frac{1}{n} \sum_{i=1}^{n} L(\hat{y_i}, y_i).
\end{equation}

The 0-1 loss $L(\cdot)$ is defined as

\begin{equation}
L(\hat{y_i}, y_i) =
\begin{cases}
0 &\text{if } \hat{y_i} = y_i \\\\
1 &\text{if } \hat{y_i} \neq y_i,
\end{cases}
\label{eqn:0-1-loss}
\end{equation}

where $y_i$ is the $i$th true class label and $\hat{y_i}$ the $i$th predicted class label, respectively. Our objective is to learn a model $h$ that has a good generalization performance. Such a model maximizes the prediction accuracy or, vice versa, minimizes the probability, $C(h)$, of making a wrong prediction:

\begin{equation}
C(h) = \underset{(\mathbf{x}, y) \sim \mathcal{D}}{\text{Pr}}[h(\mathbf{x}) \neq y].
\end{equation}

Here, $\mathcal{D}$ is the generating distribution the dataset has been drawn from, \textbf{x} is the feature vector of a training example with class label $y$.

Lastly, since this article mostly refers to the prediction accuracy (instead of the error), we define Kronecker's Delta function:

\begin{equation}
\delta \big( L(\hat{y_i}, y_i)\big) = 1 - L(\hat{y_i}, y_i),
\end{equation}

such that

\begin{equation}
\delta\big( L(\hat{y_i}, y_i)\big) = 1 \text{ if } \hat{y_i} = y_i
\end{equation}

and

\begin{equation}
\delta\big( L(\hat{y_i}, y_i)\big) = 0 \text{ if }  \hat{y_i} \neq y_i.
\end{equation}

\paragraph{Bias.}

Throughout this article, the term bias refers to the \textit{statistical bias}  (in contrast to the bias in a machine learning system). In general terms, the bias of an estimator $\hat{\beta}$ is the difference between its expected value $E[\hat{\beta}]$ and the true value of a parameter $\beta$ being estimated:

\begin{equation}
\text{Bias} = E[\hat{\beta}] - \beta.
\end{equation}

Thus, if $\text{Bias} = E[\hat{\beta}] - \beta = 0$, then $\hat{\beta}$ is an unbiased estimator of $\beta$. More concretely, we compute the prediction bias as the difference between the expected prediction accuracy of a model and its true prediction accuracy. For example, if we computed the prediction accuracy on the training set, this would be an optimistically biased estimate of the absolute accuracy of a model since it would overestimate its true accuracy.

\paragraph{Variance.} The variance is simply the statistical variance of the estimator $\hat{\beta}$ and its expected value $E[\hat{\beta]}$, for instance, the squared difference of the :

\begin{equation}
\text{Variance} = E \Big[ \big(\hat{\beta} - E[\hat{\beta}]\big)^2 \Big].
\end{equation}

The variance is a measure of the variability of a model's predictions if we repeat the learning process multiple times with small fluctuations in the training set. The more sensitive the model-building process is towards these fluctuations, the higher the variance.\footnote{For a more detailed explanation of the bias-variance decomposition of loss functions, and how high variance relates to overfitting and high bias relates to underfitting, please see my lecture notes I made available at \url{https://github.com/rasbt/stat479-machine-learning-fs18/blob/master/08_eval-intro/08_eval-intro_notes.pdf}.}

Finally, let us disambiguate the terms model, hypothesis, classifier, learning algorithms, and parameters:

\paragraph{Target function.}  In predictive modeling, we are typically interested in modeling a particular process; we want to learn or approximate a specific, unknown function. The target function $f(x) = y$ is the true function $f(\cdot)$ that we want to model.

\paragraph{Hypothesis.} A hypothesis is a certain function that we believe (or hope) is similar to the true function, the target function $f(\cdot)$ that we want to model. In context of \textit{spam} classification, it would be a classification rule we came up with that allows us to separate spam from non-spam emails.

\paragraph{Model.} In the machine learning field, the terms \textit{hypothesis} and \textit{model} are often used interchangeably. In other sciences, these terms can have different meanings: A hypothesis could be the "educated guess" by the scientist, and the model would be the manifestation of this guess to test this hypothesis.

\paragraph{Learning algorithm.} Again, our goal is to find or approximate the target function, and the learning algorithm is a set of instructions that tried to model the target function using a training dataset. A learning algorithm comes with a hypothesis space, the set of possible hypotheses it can explore to model the unknown target function by formulating the final hypothesis.

\paragraph{Hyperparameters.} Hyperparameters are the \textit{tuning parameters} of a machine learning algorithm -- for example, the regularization strength of an L2 penalty in the loss function of logistic regression, or a value for setting the maximum depth of a decision tree classifier. In contrast, model parameters are the parameters that a learning algorithm fits to the training data -- the parameters of the model itself. For example, the weight coefficients (or slope) of a linear regression line and its bias term (\textit{here:} y-axis intercept) are model parameters.

\subsection{Resubstitution Validation and the Holdout Method}
\label{sec: resubstitution-validation}

The holdout method is inarguably the simplest model evaluation technique; it can be summarized as follows. First, we take a labeled dataset and split it into two parts: A training and a test set. Then, we fit a model to the training data and predict the labels of the test set. The fraction of correct predictions, which can be computed by comparing the predicted labels to the ground truth labels of the test set, constitutes our estimate of the model's prediction accuracy. Here, it is important to note that we do not want to train and evaluate a model on the same training dataset (this is called \textit{resubstitution validation} or \textit{resubstitution evaluation}), since it would typically introduce a very optimistic bias due to overfitting. In other words, we cannot tell whether the model simply memorized the training data, or whether it generalizes well to new, unseen data. (On a side note, we can estimate this so-called \textit{optimism bias} as the difference between the training and test accuracy.)

Typically, the splitting of a dataset into training and test sets is a simple process of \textit{random subsampling}. We assume that all data points have been drawn from the same probability distribution (with respect to each class). And we randomly choose ~2/3 of these samples for the training set and ~1/3 of the samples for the test set. Note that there are two problems with this approach, which we will discuss in the next sections.

\subsection{Stratification}
\label{stratification}

\begin{figure}[htb!]
 \centering
    \includegraphics[width=\linewidth]{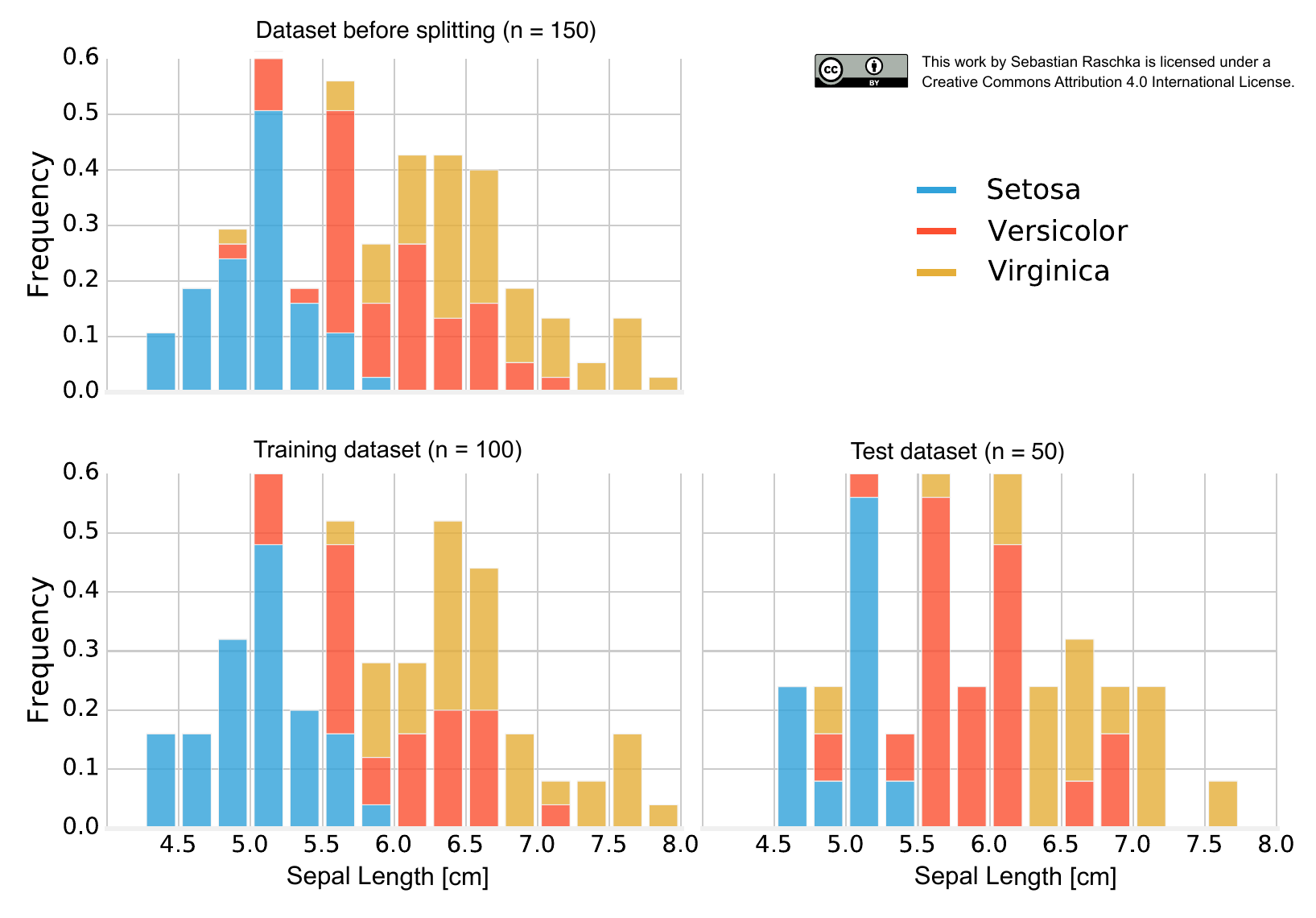}
    \caption{Distribution of \textit{Iris} flower classes upon random subsampling into training and test sets. }
    \label{fig1:iris-distribution}
\end{figure}

We have to keep in mind that a dataset represents a random sample drawn from a probability distribution, and we typically assume that this sample is representative of the true population -- more or less. Now, further subsampling without replacement alters the statistic (mean, proportion, and variance) of the sample. The degree to which subsampling without replacement affects the statistic of a sample is inversely proportional to the size of the sample. Let us have a look at an example using the \textit{Iris} dataset \footnote{https://archive.ics.uci.edu/ml/datasets/iris}, which we randomly divide into 2/3 training data and 1/3 test data as illustrated in Figure \ref{fig1:iris-distribution}. (The source code for generating this graphic is available on GitHub\footnote{https://github.com/rasbt/model-eval-article-supplementary/blob/master/code/iris-random-dist.ipynb}.)

When we randomly divide a labeled dataset into training and test sets, we violate the assumption of \textit{statistical independence}. The Iris datasets consists of 50 Setosa, 50 Versicolor, and 50 Virginica flowers; the flower species are distributed uniformly:

\begin{itemize}
 \item 33.3\% Setosa
 \item 33.3\% Versicolor
 \item 33.3\% Virginica
\end{itemize}

If a random function assigns 2/3 of the flowers (100) to the training set and 1/3 of the flowers (50) to the test set, it may yield the following (also shown in Figure \ref{fig1:iris-distribution}):

\begin{itemize}
 \item training set $\rightarrow$ 38 $\times$ Setosa, 28 $\times$ Versicolor, 34 $\times$ Virginica
 \item test set $\rightarrow$ 12 $\times$ Setosa, 22 $\times$ Versicolor, 16 $\times$ Virginica
\end{itemize}

Assuming that the Iris dataset is representative of the true population (for instance, assuming that iris flower species are distributed uniformly in nature), we just created two imbalanced datasets with non-uniform class distributions. The class ratio that the learning algorithm uses to learn the model is "38\% / 28\% / 34\%." The test dataset that is used for evaluating the model is imbalanced as well, and even worse, it is balanced in the "opposite" direction: "24\% / 44\% / 32\%." Unless the learning algorithm is completely insensitive to these perturbations, this is certainly not ideal. The problem becomes even worse if a dataset has a high class imbalance upfront, prior to the random subsampling. In the worst-case scenario, the test set may not contain any instance of a minority class at all. Thus, a recommended practice is to divide the dataset in a stratified fashion. Here, \textit{stratification} simply means that we randomly split a dataset such that each class is correctly represented in the resulting subsets (the training and the test set) -- in other words, stratification is an approach to maintain the original class proportion in resulting subsets.

It shall be noted that random subsampling in non-stratified fashion is usually not a big concern when working with relatively large and balanced datasets. However, in my opinion, stratified resampling is usually beneficial in machine learning applications. Moreover, stratified sampling is incredibly easy to implement, and Ron Kohavi provides empirical evidence \cite{kohavi1995} that stratification has a positive effect on the variance and bias of the estimate in $k$-fold cross-validation, a technique that will be discussed later in this article.

\subsection{Holdout Validation}
\label{holdout-validation}

\begin{figure}[!htb]
 \centering
    \includegraphics[width=0.9\linewidth]{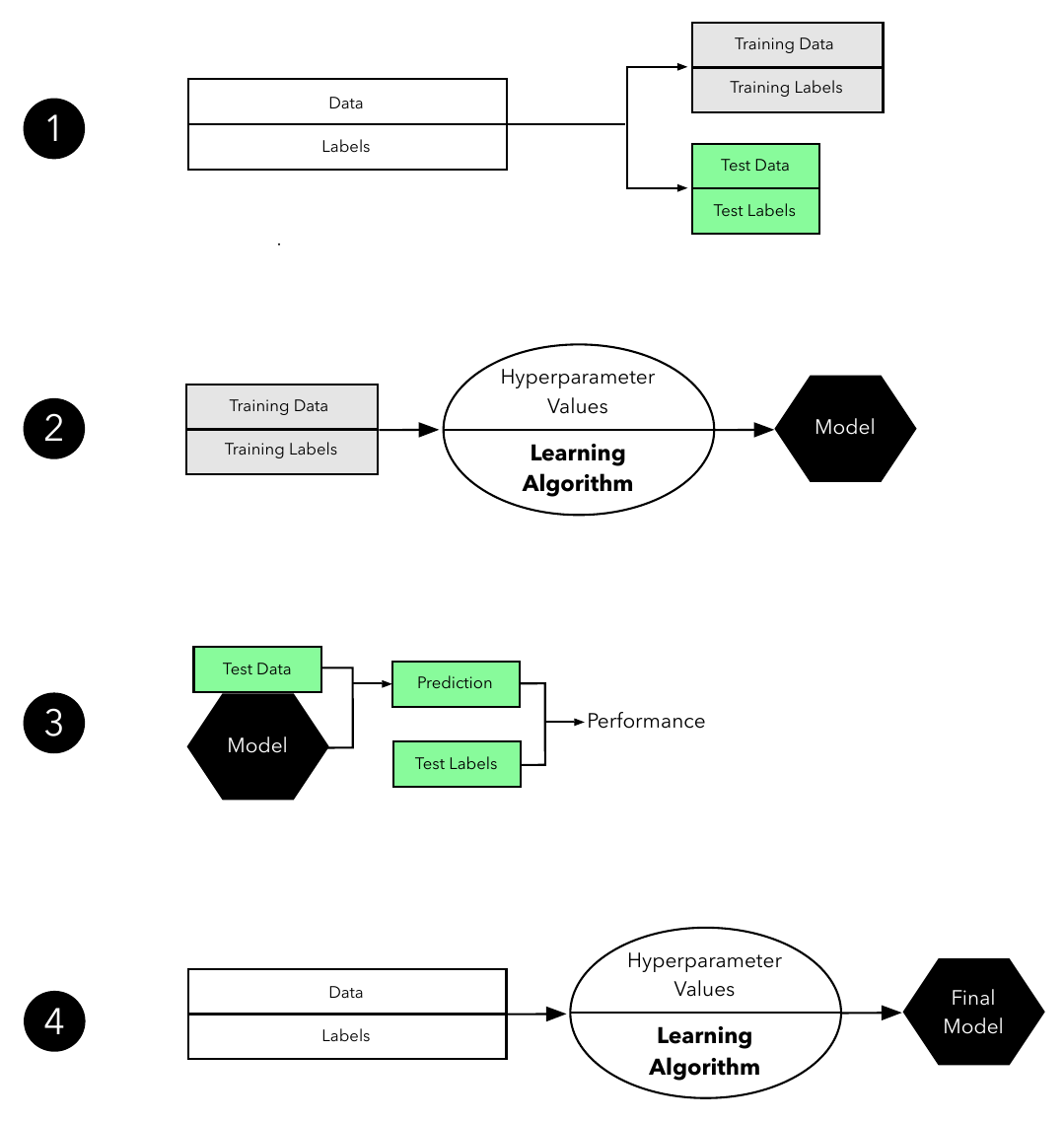}
    \caption{Visual summary of the holdout validation method.}
    \label{fig2:testing}
\end{figure}

Before diving deeper into the pros and cons of the holdout validation method, Figure \ref{fig2:testing} provides a visual summary of this method that will be discussed in the following text.

\paragraph{Step 1.} First, we randomly divide our available data into two subsets: a training and a test set. Setting test data aside is a work-around for dealing with the imperfections of a non-ideal world, such as limited data and resources, and the inability to collect more data from the generating distribution. Here, the test set shall represent new, unseen data to the model; it is important that the test set is only used once to avoid introducing bias when we estimating the generalization performance. Typically, we assign 2/3 to the training set and 1/3 of the data to the test set. Other common training/test splits are 60/40, 70/30, or 80/20 -- or even 90/10 if the dataset is relatively large.

\paragraph{Step 2.} After setting test examples aside, we pick a learning algorithm that we think could be appropriate for the given problem. As a quick reminder regarding the \textit{Hyperparameter Values} depicted in Figure \ref{fig2:testing}, hyperparameters are the parameters of our learning algorithm, or meta-parameters. And we have to specify these hyperparameter values manually -- the learning algorithm does not learn these from the training data in contrast to the actual model parameters. Since hyperparameters are not learned during model fitting, we need some sort of "extra procedure" or "external loop" to optimize these separately -- this holdout approach is ill-suited for the task. So, for now, we have to go with some fixed hyperparameter values -- we could use our intuition or the default parameters of an off-the-shelf algorithm if we are using an existing machine learning library.

\paragraph{Step 3.} After the learning algorithm fit a model in the previous step, the next question is: How "good" is the performance of the resulting model? This is where the independent test set comes into play. Since the learning algorithm has not "seen" this test set before, it should provide a relatively unbiased estimate of its performance on new, unseen data. Now, we take this test set and use the model to predict the class labels. Then, we take the predicted class labels and compare them to the "ground truth," the correct class labels, to estimate the models generalization accuracy or error.

\paragraph{Step 4.} Finally, we obtained an estimate of how well our model performs on unseen data. So, there is no reason for with-holding the test set from the algorithm any longer. Since we assume that our samples are i.i.d., there is no reason to assume the model would perform worse after feeding it all the available data. As a rule of thumb, the model will have a better generalization performance if the algorithms uses more informative data -- assuming that it has not reached its capacity, yet.

\subsection{Pessimistic Bias}

Section \ref{sec: resubstitution-validation} (Resubstitution Validation and the Holdout Method) referenced two types of problems that occur when a dataset is split into separate training and test sets. The first problem that occurs is the  violation of independence and the changing class proportions upon subsampling (discussed in Section \ref{stratification}). Walking through the holdout validation method (Section \ref{holdout-validation}) touched upon a second problem we encounter upon subsampling a dataset: Step 4 mentioned \textit{capacity} of a model, and whether additional data could be useful or not. To follow up on the capacity issue: If a model has \textit{not} reached its capacity, the performance estimate would be pessimistically biased. This assumes that the algorithm could learn a better model if it was given more data -- by splitting off a portion of the dataset for testing, we withhold valuable data for estimating the generalization performance (for instance, the test dataset). To address this issue, one might fit the model to the whole dataset after estimating the generalization performance (see Figure \ref{fig2:testing} step 4). However, using this approach, we cannot estimate its generalization performance of the refit model, since we have now "burned" the test dataset. It is a dilemma that we cannot really avoid in real-world application, but we should be aware that our estimate of the generalization performance may be pessimistically biased if only a portion of the dataset, the training dataset, is used for model fitting (this is especially affects models fit to relatively small datasets).

\subsection{Confidence Intervals via Normal Approximation}

Using the holdout method as described in Section \ref{holdout-validation}, we computed a point estimate of the generalization performance of a model. Certainly, a confidence interval around this estimate would not only be more informative and desirable in certain applications, but our point estimate could be quite sensitive to the particular training/test split (for instance, suffering from high variance). A simple approach for computing confidence intervals of the predictive accuracy or error of a model is via the so-called \textit{normal approximation}. Here, we assume that the predictions follow a normal distribution, to compute the confidence interval on the mean on a single training-test split under the central limit theorem. The following text illustrates how this works.

As discussed earlier, we compute the prediction accuracy on a dataset $S$ (\textit{here:} test set) of size $n$ as follows:

\begin{equation}
ACC_S = \frac{1}{n} \sum_{i=1}^{n} \delta(L(\hat{y_i}, y_i)),
\end{equation}

where $L(\cdot)$ is the 0-1 loss function (Equation \ref{eqn:0-1-loss}), and $n$ denotes the number of samples in the test dataset. Further, let $\hat{y}_i$ be the predicted class label and $y_i$ be the ground truth class label of the $i$th test example, respectively.  So, we could now consider each prediction as a Bernoulli trial, and the number of correct predictions $X$ is following a binomial distribution $X \sim (n,p)$ with $n$ test examples, $k$ trials, and the probability of success $p$, where $n \in \mathbb{N} \text{ and } p \in [0,1]:$

\begin{equation}
f(k;n,p) = \Pr(X = k) = \binom n k  p^k(1-p)^{n-k},
\end{equation}

for $k = 0, 1, 2, ..., n$, where

\begin{equation}
\binom n k =\frac{n!}{k!(n-k)!}.
\end{equation}

(Here, $p$ is the probability of success, and consequently, $(1-p)$  is the probability of failure -- a wrong prediction.)

Now, the expected number of successes is computed as $\mu = np$, or more concretely, if the model has a $50\%$ success rate, we expect 20 out of 40 predictions to be correct. The estimate has a variance of 

\begin{equation}
\sigma^2 = np(1-p)=10
\end{equation}

and a standard deviation of

\begin{equation}
\sigma = \sqrt{np(1-p)} = 3.16.
\end{equation}

Since we are interested in the \textit{average} number of successes, not its absolute value, we compute the variance of the accuracy estimate as

\begin{equation}
\sigma^2 = \frac{1}{n} ACC_S (1 - ACC_S),
\end{equation}

and the respective standard deviation as

\begin{equation}
\sigma = \sqrt{ \frac{1}{n} ACC_S (1 - ACC_S)}.
\end{equation}

Under the normal approximation, we can then compute the confidence interval as

\begin{equation}
ACC_S \pm z \sqrt{\frac{1}{n}ACC_S \left(1 - ACC_S \right)},
\end{equation}

where $\alpha$ is the error quantile and $z$ is the $1- \frac{\alpha}{2}$ quantile of a standard normal distribution. For a typical confidence interval of 95\%, ($\alpha=0.05$), we have $z=1.96$.

In practice, however, I would rather recommend repeating the training-test split multiple times to compute the confidence interval on the mean estimate (for instance, averaging the individual runs). In any case, one interesting take-away for now is that having fewer samples in the test set increases the variance (see $n$ in the denominator above) and thus widens the confidence interval. Confidence intervals and estimating uncertainty will be discussed in more detail in the next section, Section \ref{sec:boostrapping-and-uncertainties}.

\section{Bootstrapping and Uncertainties}
\label{sec:boostrapping-and-uncertainties}

\subsection{Overview}

The previous section (Section \ref{sec:introduction}, Introduction: Essential Model Evaluation Terms and Techniques) introduced the general ideas behind model evaluation in supervised machine learning. We discussed the holdout method, which helps us to deal with real world limitations such as limited access to new, labeled data for model evaluation. Using the holdout method, we split our dataset into two parts: A training and a test set. First, we provide the training data to a supervised learning algorithm. The learning algorithm builds a model from the training set of labeled observations. Then, we evaluate the predictive performance of the model on an independent test set that shall represent new, unseen data. Also, we briefly introduced the normal approximation, which requires us to make certain assumptions that allow us to compute confidence intervals for modeling the uncertainty of our performance estimate based on a single test set, which we have to take with a grain of salt.

This section introduces some of the advanced techniques for model evaluation. We will start by discussing techniques for estimating the uncertainty of our estimated model performance as well as the model's variance and stability. And after getting these basics under our belt, we will look at cross-validation techniques for model selection in the next article in this series. As we remember from Section \ref{sec:introduction}, there are three related, yet different tasks or reasons why we care about model evaluation:

\begin{enumerate}
\item We want to estimate the generalization accuracy, the predictive performance of a model on future (unseen) data.
\item We want to increase the predictive performance by tweaking the learning algorithm and selecting the best-performing model from a given hypothesis space.
\item We want to identify the machine learning algorithm that is best-suited for the problem at hand. Hence, we want to compare different algorithms, selecting the best-performing one as well as the best-performing model from the algorithm's hypothesis space.
\end{enumerate}

(The code for generating the figures discussed in this section are available on GitHub\footnote{https://github.com/rasbt/model-eval-article-supplementary/blob/master/code/resampling-and-kfold.ipynb}.)

\subsection{Resampling}

The first section of this article introduced the prediction accuracy or error measures of classification models. To compute the classification error or accuracy on a dataset $S$, we defined the following equation:

\begin{equation}
\text{ERR}_S = \frac{1}{n} \sum_{i=1}^{n} L\big(\hat{y}_i, y_i \big) = 1 - \text{ACC}_S.
\end{equation}

Here, $L(\cdot)$ represents the 0-1 loss, which is computed from a predicted class label ($\hat{y}_i$) and a true class label ($y_i$) for $i=1, ..., n$ in dataset $S$:

\begin{equation}
    L(\hat{y_i}, y_i) =
    \begin{cases}
    0 &\text{if } \hat{y_i} = y_i \\\\
    1 &\text{if } \hat{y_i} \neq y_i.
    \end{cases}
\end{equation}

In essence, the classification error is simply the count of incorrect predictions divided by the number of samples in the dataset. Vice versa, we compute the prediction accuracy as the number of correct predictions divided by the number of samples.

Note that the concepts presented in this section also apply to other types of supervised learning, such as regression analysis. To use the resampling methods presented in the following sections for regression models, we swap the accuracy or error computation by, for example, the mean squared error (MSE):

\begin{equation}
\text{MSE}_S = \frac{1}{n} \sum_{i=1}^{n}  ( \hat{y}_i - y_i)^2.
\end{equation}

\begin{figure}[htb!]
 \centering
    \includegraphics[width=0.7\linewidth]{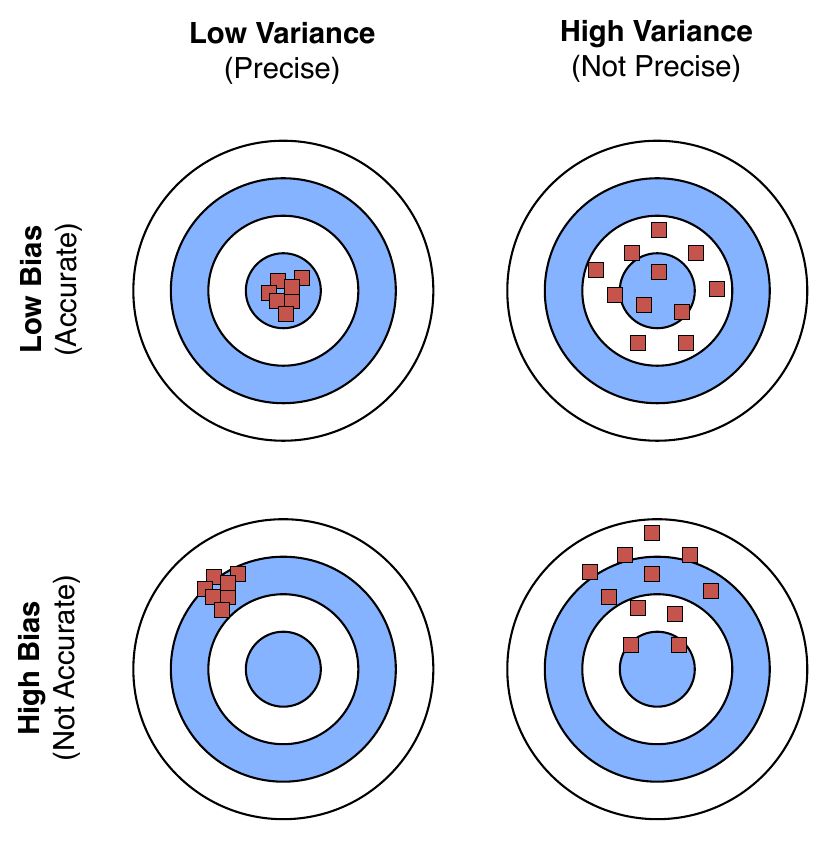}
    \caption{Illustration of bias and variance.}
    \label{fig3:bias-variance}
\end{figure}

As discussed in Section \ref{sec:introduction}, performance estimates may suffer from bias and variance, and we are interested in finding a good trade-off. For instance, the resubstitution evaluation (fitting a model to a training set and using the same training set for model evaluation) is heavily optimistically biased. Vice versa, withholding a large portion of the dataset as a test set may lead to pessimistically biased estimates. While reducing the size of the test set may decrease this pessimistic bias, the variance of a performance estimates will most likely increase. An intuitive illustration of the relationship between bias and variance is given in Figure \ref{fig3:bias-variance}. This section will introduce alternative resampling methods for finding a good balance between bias and variance for model evaluation and selection.

\begin{figure}[htb!]
 \centering
    \includegraphics[width=0.7\linewidth]{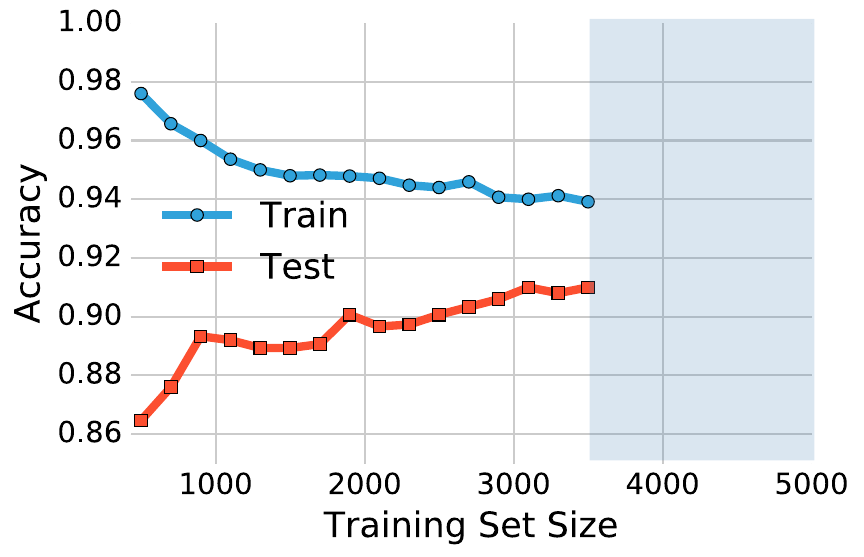}
    \caption{Learning curves of softmax classifiers fit to MNIST.}
    \label{fig4:softmax-mnist}
\end{figure}

The reason why a proportionally large test sets increase the pessimistic bias is that the model may not have reached its full capacity, yet. In other words, the learning algorithm could have formulated a more powerful, more generalizable hypothesis for classification if it had seen more data. To demonstrate this effect, Figure \ref{fig4:softmax-mnist} shows learning curves of a softmax classifiers, which were fitted to small subsets of the MNIST\footnote{http://yann.lecun.com/exdb/mnist} dataset.

To generate the learning curves shown in Figure  \ref{fig4:softmax-mnist}, 500 random samples of each of the ten classes from MNIST -- instances of the handwritten digits 0 to 9 -- were drawn. The 5000-sample MNIST subset was then randomly divided into a 3500-sample training subset and a test set containing 1500 samples while keeping the class proportions intact via stratification. Finally, even smaller subsets of the 3500-sample training set were produced via randomized, stratified splits, and these subsets were used to fit softmax classifiers and the same 1500-sample test set was used to evaluate their performances (samples may overlap between these training subsets). Looking at the plot above, we can see two distinct trends. First, the resubstitution accuracy (training set) declines as the number of training samples grows. Second, we observe an improving generalization accuracy (test set) with an increasing training set size. These trends can likely be attributed to a reduction in overfitting. If the training set is small, the algorithm is more likely picking up noise in the training set so that the model fails to generalize well to data that it has not seen before. This observation also explains the pessimistic bias of the holdout method: A training algorithm may benefit from more training data, data that was withheld for testing. Thus, after we evaluated a model, we may want to run the learning algorithm once again on the complete dataset before we use it in a real-world application.

\begin{figure}[!htb]
 \centering
    \includegraphics[width=1.0\linewidth]{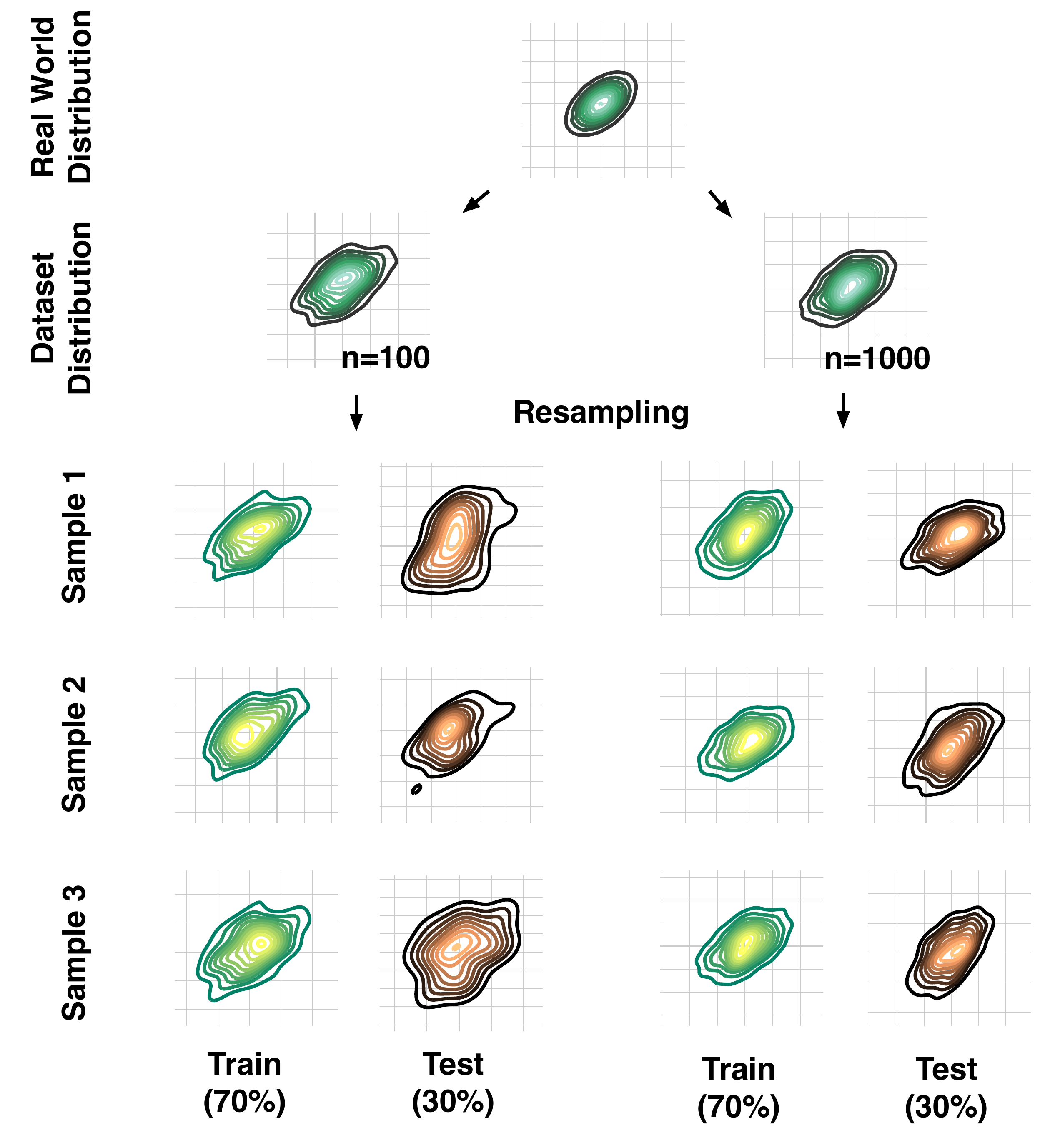}
    \caption{Repeated subsampling from a two-dimensional Gaussian distribution.}
    \label{fig5:resampling-gauss}
\end{figure}

Now, that we established the point of pessimistic biases for disproportionally large test sets, we may ask whether it is a good idea to decrease the size of the test set. Decreasing the size of the test set brings up another problem: It may result in a substantial variance of a model's performance estimate. The reason is that it depends on which instances end up in training set, and which particular instances end up in test set. Keeping in mind that each time we resample a dataset, we alter the statistics of the distribution of the sample. Most supervised learning algorithms for classification and regression as well as the performance estimates operate under the assumption that a dataset is representative of the population that this dataset sample has been drawn from. As discussed in Section \ref{stratification}, stratification helps with keeping the sample proportions intact upon splitting a dataset. However, the change in the underlying sample statistics along the features axes is still a problem that becomes more pronounced if we work with small datasets, which is illustrated in Figure \ref{fig5:resampling-gauss}. 

\subsection{Repeated Holdout Validation}

One way to obtain a more robust performance estimate that is less variant to how we split the data into training and test sets is to repeat the holdout method $k$ times with different random seeds and compute the average performance over these $k$ repetitions:

\begin{equation}
\text{ACC}_{avg} = \frac{1}{k} \sum^{k}_{j=1} {ACC_j},
\end{equation}

where $\text{ACC}_j$ is the accuracy estimate of the $j$th test set of size $m$,

\begin{equation}
\text{ACC}_j = 1 - \frac{1}{m} \sum_{i=1}^{m} L\big(\hat{y}_i, y_i \big).
\end{equation}

\begin{figure}[!htb]
 \centering
    \includegraphics[width=1.0\linewidth]{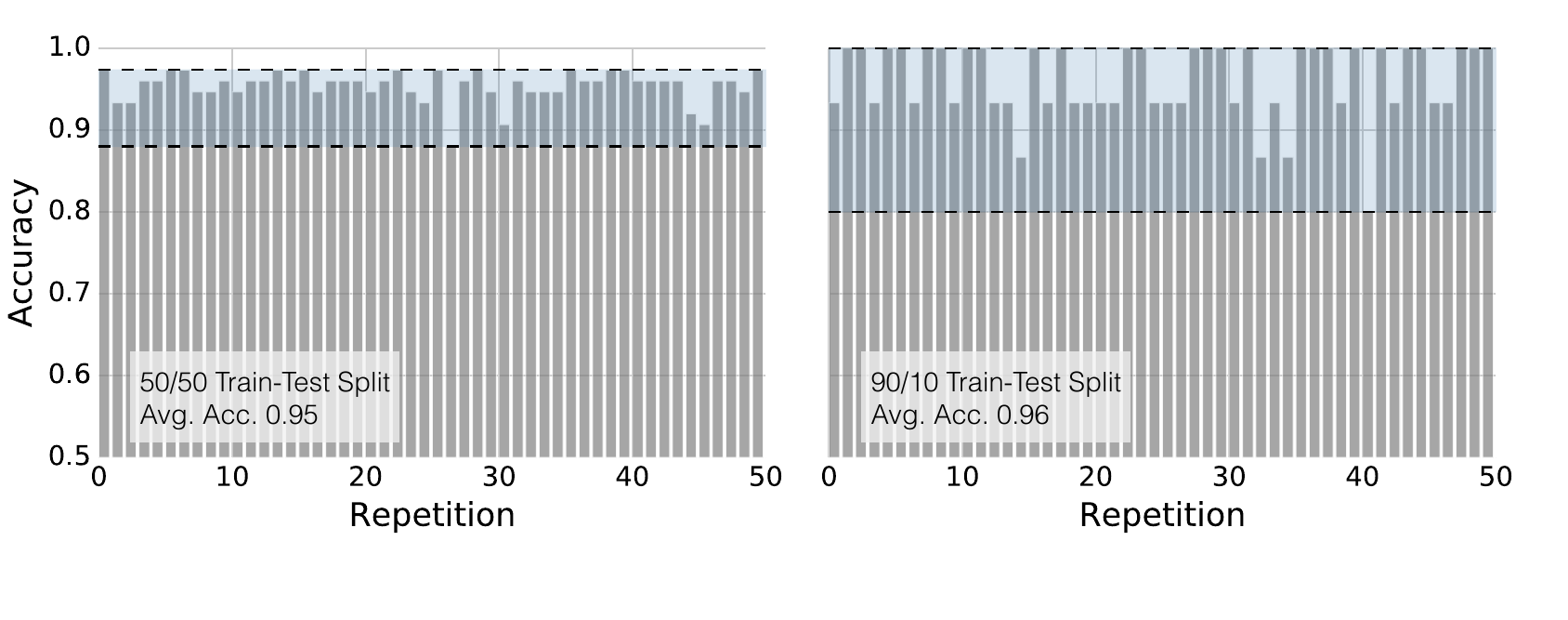}
    \caption{Repeated holdout validation with 3-nearest neighbor classifiers fit to the Iris dataset.}
    \label{fig6:model-eval-iris}
\end{figure}

This repeated holdout procedure, sometimes also called Monte Carlo Cross-Validation, provides a better estimate of how well our model may perform on a random test set, compared to the standard holdout validation method. Also, it provides information about the model's stability -- how the model, produced by a learning algorithm, changes with different training set splits. Figure  \ref{fig6:model-eval-iris} shall illustrate how repeated holdout validation may look like for different training-test split using the Iris dataset to fit to 3-nearest neighbors classifiers.

The left subplot in Figure \ref{fig6:model-eval-iris} was generated by performing  50 stratified training/test splits with 75 samples in the test and training set each; a 3-nearest neighbors model was fit to the training set and evaluated on the test set in each repetition. The average accuracy of these 50 50/50 splits was 95\%. The same procedure was used to produce the right subplot in Figure \ref{fig6:model-eval-iris}. Here, the test sets consisted of only 15 samples each due to the 90/10 splits, and  the average accuracy over these 50 splits was 96\%. Figure \ref{fig6:model-eval-iris} demonstrates two of the points that were previously discussed. First, we see that the variance of our estimate increases as the size of the test set decreases. Second, we see a small increase in the pessimistic bias when we decrease the size of the training set -- we withhold more training data in the 50/50 split, which may be the reason why the average performance over the 50 splits is slightly lower compared to the 90/10 splits.

The next section introduces an alternative method for evaluating a model's performance; it will discuss about different flavors of the bootstrap method that are commonly used to infer the uncertainty of a performance estimate.

\subsection{The Bootstrap Method and Empirical Confidence Intervals}

The previous examples of Monte Carlo Cross-Validation may have convinced us that repeated holdout validation could provide us with a more robust estimate of a model's performance on random test sets compared to an evaluation based on a single train/test split via holdout validation (Section \ref{holdout-validation}). In addition, the repeated holdout may give us an idea about the stability of our model. This section explores an alternative approach to model evaluation and for estimating uncertainty using the bootstrap method.

Let us assume that we would like to compute a confidence interval around a performance estimate to judge its certainty -- or uncertainty. How can we achieve this if our sample has been drawn from an unknown distribution? Maybe we could use the sample mean as a point estimate of the population mean, but how would we compute the variance or confidence intervals around the mean if its distribution is unknown? Sure, we could collect multiple, independent samples; this is a luxury we often do not have in real world applications, though. Now, the idea behind the bootstrap is to generate "new samples" by sampling from an empirical distribution. As a side note, the term "bootstrap" likely originated from the phrase "to pull oneself up by one's bootstraps:"

\begin{displayquote}
Circa 1900, to pull (oneself) up by (one's) bootstraps was used figuratively of an impossible task (Among the "practical questions" at the end of chapter one of Steele's "Popular Physics" schoolbook (1888) is "Why can not a man lift himself by pulling up on his boot-straps?". By 1916 its meaning expanded to include "better oneself by rigorous, unaided effort." The meaning "fixed sequence of instructions to load the operating system of a computer" (1953) is from the notion of the first-loaded program pulling itself, and the rest, up by the bootstrap. 

[Source: Online Etymology Dictionary\footnote{https://www.etymonline.com/word/bootstrap}]
\end{displayquote}

The bootstrap method is a resampling technique for estimating a sampling distribution, and in the context of this article, we are particularly interested in estimating the uncertainty of a performance estimate -- the prediction accuracy or error. The bootstrap method was introduced by Bradley Efron in 1979 \cite{efron1992bootstrap}. About 15 years later, Bradley Efron and Robert Tibshirani even devoted a whole book to the bootstrap, "An Introduction to the Bootstrap" \cite{efron1994introduction}, which is a highly recommended read for everyone who is interested in more details on this topic. In brief, the idea of the bootstrap method is to generate new data from a population by repeated sampling from the original dataset with replacement -- in contrast, the repeated holdout method can be understood as sampling without replacement. Walking through it step by step, the bootstrap method works like this:

\begin{enumerate}
\item We are given a dataset of size $n$.
\item For $b$ bootstrap rounds:
\subitem We draw one single instance from this dataset and assign it to the $j$th bootstrap sample. We repeat this step until our bootstrap sample has size n -- the size of the original dataset. Each time, we draw samples from the same original dataset such that certain examples may appear more than once in a bootstrap sample and some not at all.
\item We fit a model to each of the $b$ bootstrap samples and compute the resubstitution accuracy.
\item We compute the model accuracy as the average over the $b$ accuracy estimates (Equation \ref{eqn:boot-acc}).
\end{enumerate}

\begin{equation}
\text{ACC}_{boot} = \frac{1}{b} \sum_{j=1}^b \frac{1}{n} \sum_{i=1}^{n} \bigg( 1 - L\big(\hat{y}_i, y_i \big) \bigg)
\label{eqn:boot-acc}
\end{equation}

\begin{figure}[!htb]
 \centering
    \includegraphics[width=1.0\linewidth]{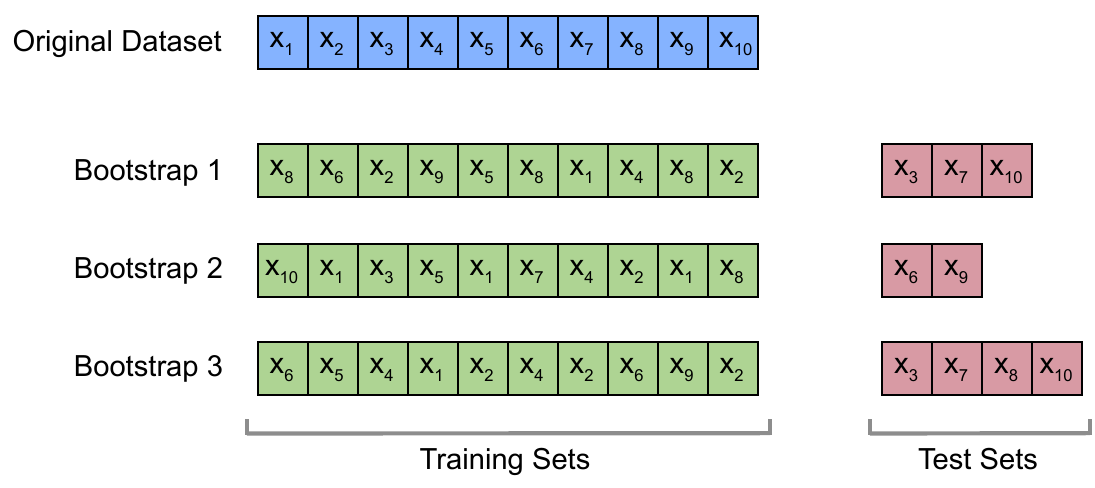}
    \caption{Illustration of training and test data splits in the Leave-One-Out Bootstrap (LOOB).}
    \label{fig7:loob}
\end{figure}

As discussed previously, the resubstitution accuracy usually leads to an extremely optimistic bias, since a model can be overly sensible to noise in a dataset. Originally, the bootstrap method aims to determine the statistical properties of an estimator when the underlying distribution was unknown and additional samples are not available. So, in order to exploit this method for the evaluation of predictive models, such as hypotheses for classification and regression, we may prefer a slightly different approach to bootstrapping using the so-called Leave-One-Out Bootstrap (LOOB) technique. Here, we use out-of-bag samples as test sets for evaluation instead of evaluating the model on the training data. Out-of-bag samples are the unique sets of instances that are not used for model fitting as shown in Figure \ref{fig7:loob}.

Figure \ref{fig7:loob} illustrates how three random bootstrap samples drawn from an exemplary ten-sample dataset ($x_1, x_2, ..., x_10$) and how the out-of-bag sample might look like. In practice, Bradley Efron and Robert Tibshirani recommend drawing 50 to 200 bootstrap samples as being sufficient for producing reliable estimates \cite{efron1994introduction}.

Taking a step back, let us assume that a sample that has been drawn from a normal distribution. Using basic concepts from statistics, we use the sample mean $\bar{x}$ as a point estimate of the population mean $\mu$:

\begin{equation}
\bar{x} = \frac{1}{n} \sum_{i=1}^{n} x_i.
\end{equation}

Similarly, the variance $\sigma^2$ is estimated as follows:

\begin{equation}
\text{VAR} = \frac{1}{n-1} \sum_{i=1}^{n} (x_i - \bar{x})^2.
\end{equation}

Consequently, the standard error (SE) is computed as the standard deviation's estimate ($\text{SD} \approx \sigma$) divided by the square root of the sample size:

\begin{equation}
\text{SE} = \frac{\text{SD}}{\sqrt{n}}.
\end{equation}

Using the standard error we can then compute a 95\% confidence interval of the mean according to

\begin{equation}
\bar{x} \pm z \times \frac{\sigma}{\sqrt{n}},
\end{equation}

such that

\begin{equation}
\bar{x} \pm t \times \text{SE},
\end{equation}

with $z=1.96$ for the 95 \% confidence interval. Since SD is the standard deviation of the population ($\sigma$) estimated from the sample, we have to consult a $t$-table to look up the actual value of $t$, which depends on the size of the sample -- or the \textit{degrees of freedom} to be precise. For instance, given a sample with $n=100$, we find that $t_{95} = 1.984$.

Similarly, we can compute the 95\% confidence interval of the bootstrap estimate starting with the mean accuracy,

\begin{equation}
\text{ACC}_{boot} = \frac{1}{b} \sum_{i=1}^{b} \text{ACC}_i,
\end{equation}

and use it to compute the standard error

\begin{equation}
\text{SE}_{boot} = \sqrt{ \frac{1}{b-1} \sum_{i=1}^{b} (\text{ACC}_i - \text{ACC}_{boot})^2 }.
\end{equation}

Here, $\text{ACC}_i$ is the value of the statistic (the estimate of $ACC$) calculated on the $i$th bootstrap replicate. And the standard deviation of the values $\text{ACC}_1, \text{ACC}_1, ..., \text{ACC}_b$ is the estimate of the standard error of ACC \cite{efron1994introduction}.

Finally, we can then compute the confidence interval around the mean estimate as

\begin{equation}
\text{ACC}_{boot} \pm t \times \text{SE}_{boot}.
\end{equation}

Although the approach outlined above seems intuitive, what can we do if our samples do not follow a normal distribution? A more robust, yet computationally straight-forward approach is the percentile method as described by B. Efron \cite{efron1981nonparametric}. Here, we pick the lower and upper confidence bounds as follows:

\begin{itemize}
 \item $\text{ACC}_{lower} = \alpha_{1}$th percentile of the $\text{ACC}_{boot}$ distribution
 \item $\text{ACC}_{upper} = \alpha_{2}$th percentile of the $\text{ACC}_{boot}$ distribution,
\end{itemize}

where $\alpha_1 = \alpha$ and $\alpha_2=1-\alpha$, and $\alpha$ is the degree of confidence for computing the $100 \times (1 - 2 \times \alpha)$ confidence interval. For instance, to compute a 95\% confidence interval, we pick $\alpha=0.025$ to obtain the the 2.5th and 97.5th percentiles of the $b$ bootstrap samples distribution as our upper and lower confidence bounds.

\begin{figure}[!htb]
 \centering
    \includegraphics[width=1.0\linewidth]{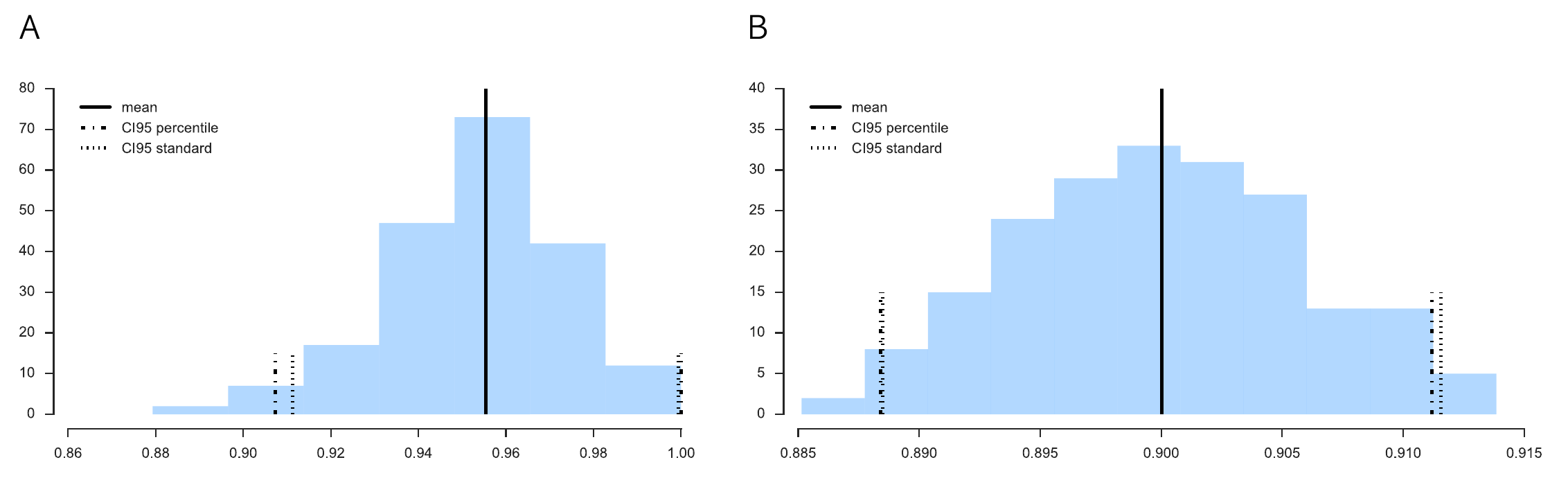}
    \caption{Comparison of the standard and percentile method for computing confidence intervals from leave-one-out bootstrap samples. Subpanel A evaluates 3-nearest neighbors models on Iris, and sublpanel B shows the results of softmax regression models on MNIST.}
    \label{fig8:bootstrap-histo}
\end{figure}

In practice, if the data is indeed (roughly) following a normal distribution, the "standard" confidence interval and percentile method typically agree as illustrated in the Figure \ref{fig8:bootstrap-histo}.

In 1983, Bradley Efron described the \textit{.632 Estimate}, a further improvement to address the pessimistic bias of the bootstrap cross-validation approach described above \cite{efron1983estimating}. The pessimistic bias in the "classic" bootstrap method can be attributed to the fact that the bootstrap samples only contain approximately 63.2\% of the unique examples from the original dataset. For instance, we can compute the probability that a given example from a dataset of size $n$ is not drawn as a bootstrap sample as follows:

\begin{equation}
P (\text{not chosen}) =  \bigg(1 - \frac{1}{n}\bigg)^n,
\end{equation}

which is asymptotically equivalent to $\frac{1}{e} \approx 0.368$ as $n \rightarrow \infty$.

\begin{figure}[!htb]
 \centering
    \includegraphics[width=0.8\linewidth]{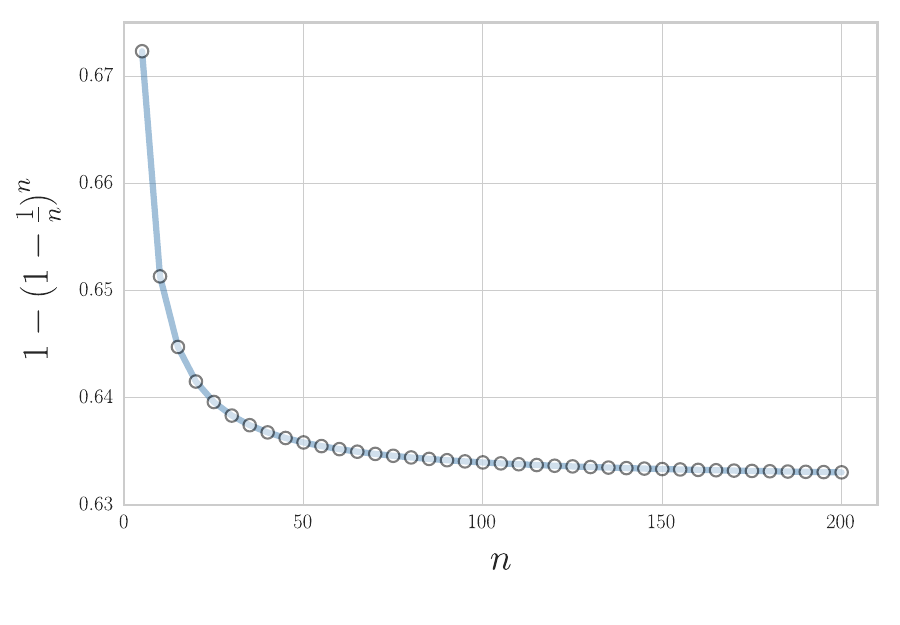}
    \caption{Probability of including an example from the dataset in a bootstrap sample for different dataset sizes $n$.}
    \label{fig9:bootstrap-prob}
\end{figure}

Vice versa, we can then compute the probability that a sample \textit{is} chosen as

\begin{equation}
P (\text{chosen}) = 1 - \bigg(1 - \frac{1}{n}\bigg)^n \approx 0.632
\end{equation}

for reasonably large datasets, so that we select approximately $0.632 \times n$ unique examples as bootstrap training sets and reserve $0.382 \times n$ out-of-bag examples for testing in each iteration, which is illustrated in Figure \ref{fig9:bootstrap-prob}.

Now, to address the bias that is due to this the sampling with replacement, Bradley Efron proposed the \textit{.632 Estimate} mentioned earlier, which is computed via the following equation:

\begin{equation}
\text{ACC}_{boot} = \frac{1}{b} \sum_{i=1}^b \big(0.632 \cdot \text{ACC}_{h, i} + 0.368 \cdot \text{ACC}_{r, i}\big),
\end{equation}

where $\text{ACC}_{r,i}$ is the resubstitution accuracy, and $\text{ACC}_{h,i}$ is the accuracy on the out-of-bag sample. Now, while the .632 Boostrap attempts to address the pessimistic bias of the estimate, an optimistic bias may occur with models that tend to overfit so that Bradley Efron and Robert Tibshirani proposed \textit{The .632+ Bootstrap Method} \cite{efron1997improvements}. Instead of using a fixed weight $\omega=0.632$ in 

\begin{equation}
ACC_{\text{boot}} = \frac{1}{b} \sum_{i=1}^b \big(\omega \cdot \text{ACC}_{h, i} + (1-\omega) \cdot \text{ACC}_{r, i} \big),
\label{eqn:acc-boot}
\end{equation}

we compute the weight $\omega$ as 

\begin{equation}
\omega = \frac{0.632}{1 - 0.368 \times R},
\end{equation}

where \textit{R} is the \textit{relative overfitting rate}:

\begin{equation}
R = \frac{(-1) \times (\text{ACC}_{h, i} - \text{ACC}_{r, i})}{\gamma - (1 -\text{ACC}_{h, i})}.
\end{equation}

(Since we are plugging $\omega$ into Equation \ref{eqn:acc-boot} for computing $\text{ACC}_boot$ that we defined above, $\text{ACC}_{h,i}$ and $\text{ACC}_{r,i}$ still refer to the resubstitution and out-of-bag accuracy estimates in the $i$th bootstrap round, respectively.)

Further, we need to determine the \textit{no-information rate} $\gamma$ in order to compute $R$. For instance, we can compute $\gamma$ by fitting a model to a dataset that contains all possible combinations between samples $x'_i$ and target class labels $y_i$ -- we pretend that the observations and class labels are independent: 

\begin{equation}
\gamma = \frac{1}{n^2} \sum_{i=1}^{n} \sum_{i '=1}^{n} L(y_{i}, f(x_{i '})).
\end{equation}

Alternatively, we can estimate the no-information rate $\gamma$ as follows:

\begin{equation}
\gamma = \sum_{k=1}^K p_k (1 - q_k),
\end{equation}

where $p_k$ is the proportion of class $k$ examples observed in the dataset, and $q_k$ is the proportion of class $k$ examples that the classifier predicts in the dataset.

The OOB bootstrap, 0.632 bootstrap, and .632+ boostrap method discussed in this section are implemented in MLxtend \cite{raschka2018mlxtend} to enable comparison studies.\footnote{\url{http://rasbt.github.io/mlxtend/user_guide/evaluate/bootstrap_point632_score/}}

This section continued the discussion around biases and variances in evaluating machine learning models in more detail. Further, it introduced the repeated hold-out method that may provide us with some further insights into a model's stability. Then, we looked at the bootstrap method and explored different flavors of this bootstrap method that help us estimate the uncertainty of our performance estimates. After covering the basics of model evaluation in this and the previous section, the next section introduces hyperparameter tuning and model selection.

\section{Cross-validation and Hyperparameter Optimization}
\label{sec:cross-val}

\subsection{Overview}

Almost every machine learning algorithm comes with a large number of settings that we, the machine learning researchers and practitioners, need to specify. These tuning knobs, the so-called hyperparameters, help us control the behavior of machine learning algorithms when optimizing for performance, finding the right balance between bias and variance. Hyperparameter tuning for performance optimization is an art in itself, and there are no hard-and-fast rules that guarantee best performance on a given dataset. The previous sections covered holdout and bootstrap techniques for estimating the generalization performance of a model. The bias-variance trade-off was discussed in the context of estimating the generalization performance as well as methods for computing the uncertainty of performance estimates. This third section focusses on different methods of cross-validation for model evaluation and model selection. It covers cross-validation techniques to rank models from several hyperparameter configurations and estimate how well these generalize to independent datasets.

(Code for generating the figures discussed in this section is available on GitHub\footnote{https://github.com/rasbt/model-eval-article-supplementary/blob/master/code/resampling-and-kfold.ipynb}.)

\subsection{About Hyperparameters and Model Selection}

Previously, the \textit{holdout method} and different flavors of the \textit{bootstrap} were introduced to estimate the generalization performance of our predictive models. We split the dataset into two parts: a training and a test dataset. After the machine learning algorithm fit a model to the training set, we evaluated it on the independent test set that we withheld from the machine learning algorithm during model fitting. While we were discussing challenges such as the bias-variance trade-off, we used fixed hyperparameter settings in our learning algorithms, such as the number of $k$ in the \textit{$k$-nearest neighbors algorithm}. We defined hyperparameters as the parameters of the learning algorithm itself, which we have to specify a priori -- before model fitting. In contrast, we referred to the parameters of our resulting model as the \textit{model parameters}.

\begin{figure}[!htb]
 \centering
    \includegraphics[width=0.65\linewidth]{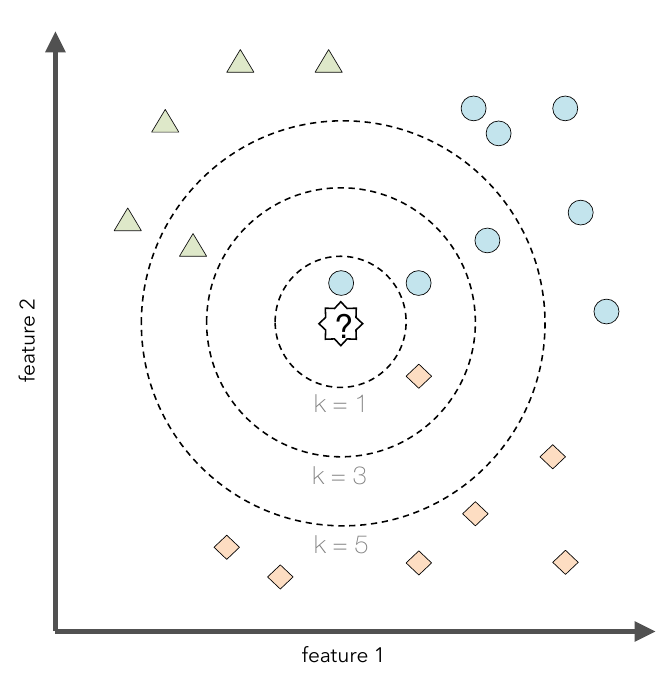}
    \caption{Illustration of the $k$-nearest neighbors algorithm with different choices for $k$.}
    \label{fig10:knn}
\end{figure}

So, what are hyperparameters, exactly? Considering the $k$-nearest neighbors algorithm, one example of a hyperparameter is the integer value of $k$ (Figure \ref{fig10:knn}). If we set $k=3$, the $k$-nearest neighbors algorithm will predict a class label based on a majority vote among the 3-nearest neighbors in the training set. The distance metric for finding these nearest neighbors is yet another hyperparameter of this algorithm.

Now, the $k$-nearest neighbors algorithm may not be an ideal choice for illustrating the difference between hyperparameters and model parameters, since it is a \textit{lazy learner} and a nonparametric method. In this context, lazy learning (or instance-based learning) means that there is no training or model fitting stage: A $k$-nearest neighbors model literally stores or memorizes the training data and uses it only at prediction time. Thus, each training instance represents a parameter in the $k$-nearest neighbors model. In short, nonparametric models are models that cannot be described by a fixed number of parameters that are being adjusted to the training set. The structure of parametric models is not decided by the training data rather than being set \textit{a priori}; nonparamtric models do not assume that the data follows certain probability distributions unlike parametric methods (exceptions of nonparametric methods that make such assumptions are Bayesian nonparametric methods). Hence, we may say that nonparametric methods make fewer assumptions about the data than parametric methods.

\begin{figure}[!htb]
 \centering
    \includegraphics[width=0.95\linewidth]{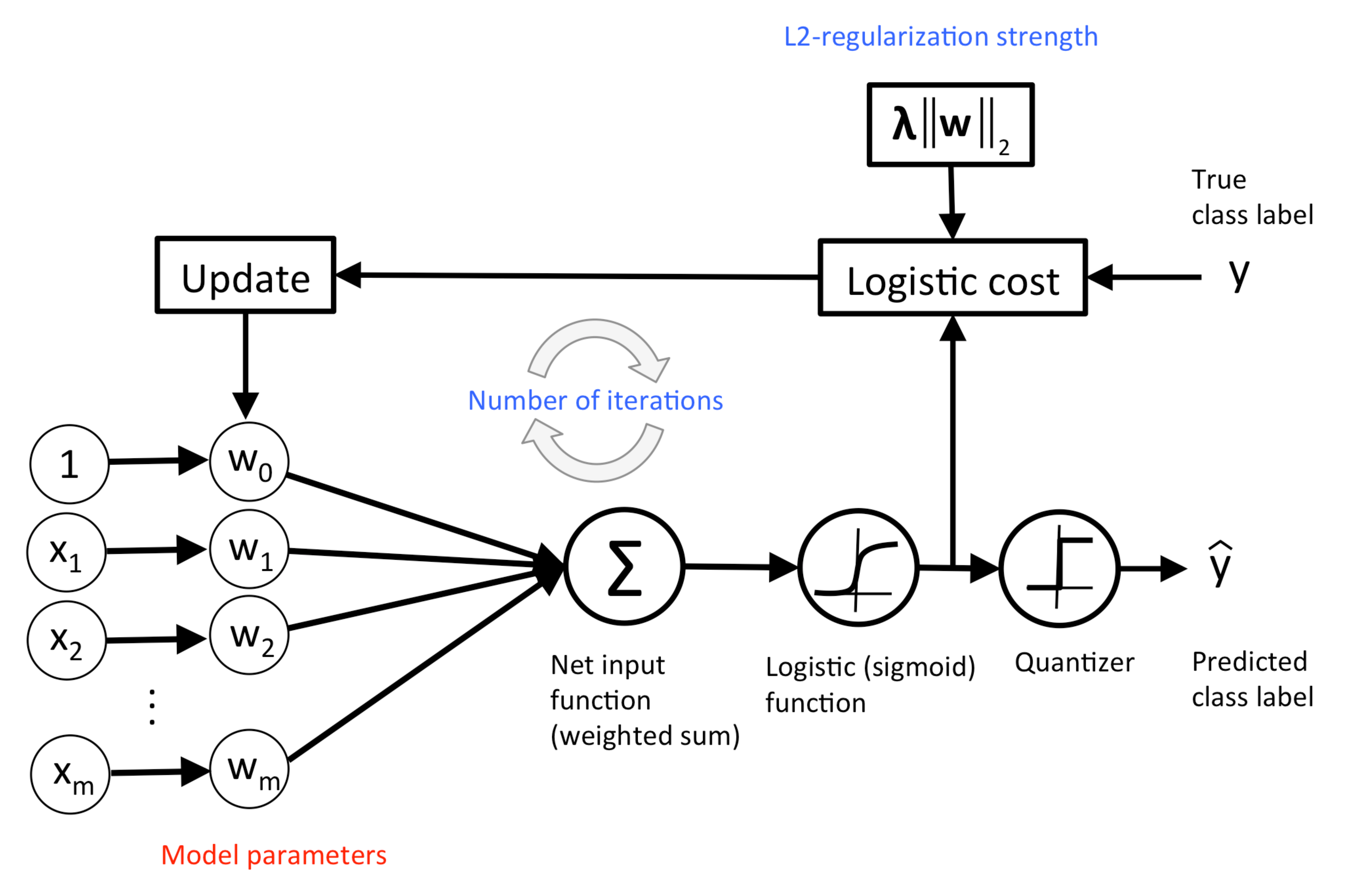}
    \caption{Conceptual overview of logistic regression.}
    \label{fig11:logistic-regression}
\end{figure}

In contrast to $k$-nearest neighbors, a simple example of a parametric method is logistic regression, a generalized linear model with a fixed number of model parameters: a weight coefficient for each feature variable in the dataset plus a bias (or intercept) unit. These weight coefficients in logistic regression, the model parameters, are updated by maximizing a log-likelihood function or minimizing the logistic cost. For fitting a model to the training data, a hyperparameter of a logistic regression algorithm could be the number of iterations or passes over the training set (epochs) in gradient-based optimization. Another example of a hyperparameter would be the value of a regularization parameter such as the lambda-term in L2-regularized logistic regression (Figure \ref{fig11:logistic-regression}).

Changing the hyperparameter values when running a learning algorithm over a training set may result in different models. The process of finding the best-performing model from a set of models that were produced by different hyperparameter settings is called \textit{model selection}. The next section introduces an extension to the holdout method that is useful when carrying out this selection process.

\subsection{The Three-Way Holdout Method for Hyperparameter Tuning}

Section \ref{sec:introduction} provided an explanation why resubstitution validation is a bad approach for estimating of the generalization performance. Since we want to know how well our model generalizes to new data, we used the holdout method to split the dataset into two parts, a training set and an independent test set. Can we use the holdout method for hyperparameter tuning? The answer is "yes." However, we have to make a slight modification to our initial approach, the "two-way" split, and split the dataset into three parts: a training, a validation, and a test set.

The process of hyperparameter tuning (or hyperparameter optimization) and model selection can be regarded as a meta-optimization task. While the learning algorithm optimizes an objective function on the training set (with exception to lazy learners), hyperparameter optimization is yet another task on top of it; here, we typically want to optimize a performance metric such as classification accuracy or the area under a Receiver Operating Characteristic curve. After the tuning stage, selecting a model based on the test set performance seems to be a reasonable approach. However, reusing the test set multiple times would introduce a bias and the final performance estimate and likely result in overly optimistic estimates of the generalization performance -- one might say that "the test set leaks information." To avoid this problem, we could use a three-way split, dividing the dataset into a training, validation, and test dataset. Having a training-validation pair for hyperparameter tuning and model selections allows us to keep the test set "independent" for model evaluation. Once more, let us recall the "3 goals" of performance estimation:

\begin{enumerate}
\item We want to estimate the generalization accuracy, the predictive performance of a model on future (unseen) data.
\item We want to increase the predictive performance by tweaking the learning algorithm and selecting the best-performing model from a given hypothesis space.
\item We want to identify the machine learning algorithm that is best-suited for the problem at hand; thus, we want to compare different algorithms, selecting the best-performing one as well as the best-performing model from the algorithm's hypothesis space.
\end{enumerate}

The "three-way holdout method" is one way to tackle points 1 and 2 (more on point 3 in Section  \ref{sec:algo-comparison}). Though, if we are only interested in point 2, selecting the best model, and do not care so much about an "unbiased" estimate of the generalization performance, we could stick to the two-way split for model selection. Thinking back of our discussion about learning curves and pessimistic biases in Section \ref{sec:boostrapping-and-uncertainties}, we noted that a machine learning algorithm often benefits from more labeled data; the smaller the dataset, the higher the pessimistic bias and the variance -- the sensitivity of a model towards the data is partitioned.

\begin{figure}[!htb]
 \centering
    \includegraphics[width=0.75\linewidth]{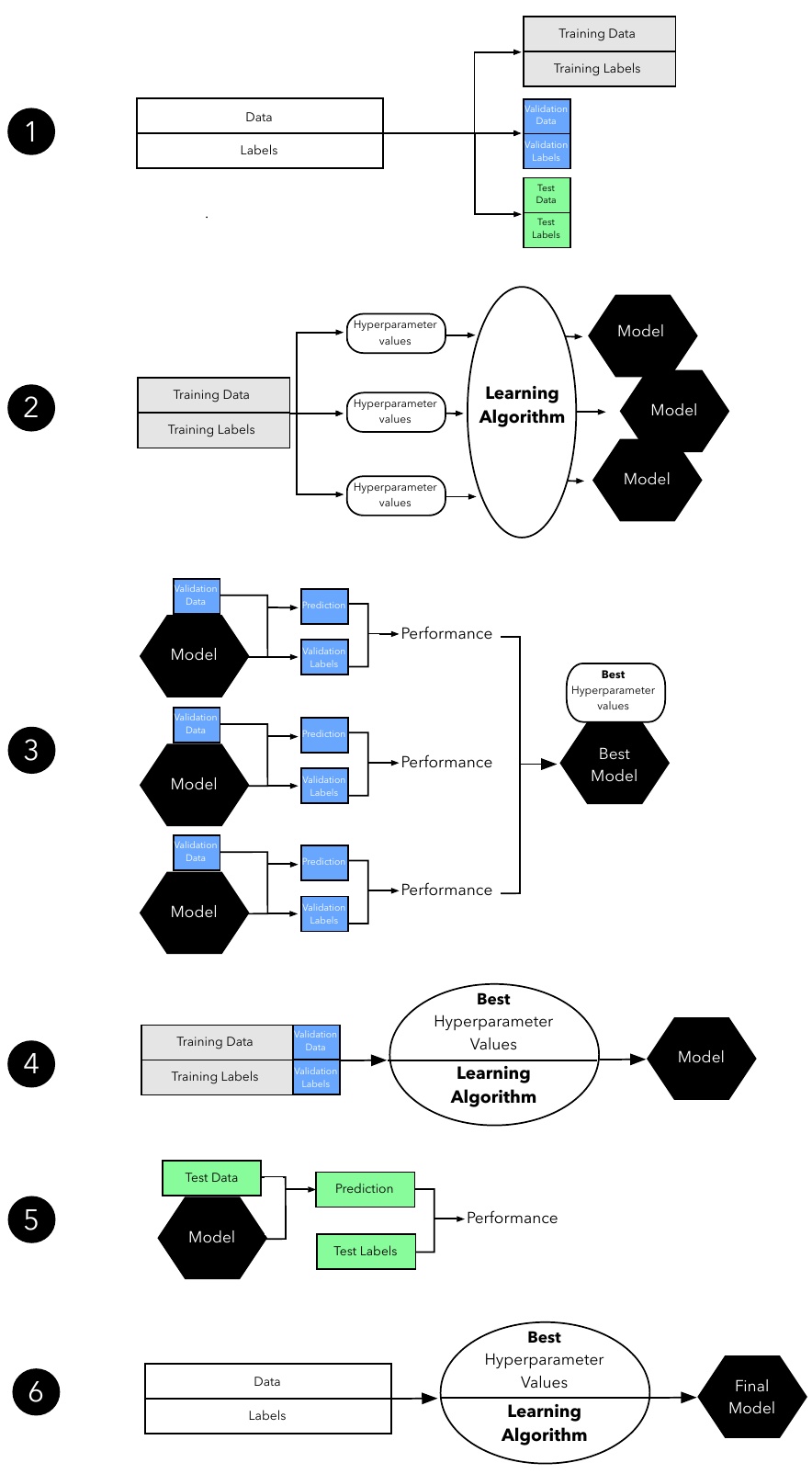}
    \caption{Illustration of the three-way holdout method for hyperparameter tuning.}
    \label{fig12:holdout-validation}
\end{figure}

"There ain't no such thing as a free lunch." The three-way holdout method for hyperparameter tuning and model selection is not the only -- and certainly often not the best -- way to approach this task. Later sections, will introduce alternative methods and discuss their advantages and trade-offs. However, before we move on to the probably most popular method for model selection, $k$-fold cross-validation (or sometimes also called "rotation estimation" in older literature), let us have a look at an illustration of the 3-way split holdout method in Figure  \ref{fig12:holdout-validation}.

Let us walk through Figure  \ref{fig12:holdout-validation} step by step.

\paragraph{Step 1.} We start by splitting our dataset into three parts, a training set for model fitting, a validation set for model selection, and a test set for the final evaluation of the selected model.

\paragraph{Step 2.} This step illustrates the hyperparameter tuning stage. We use the learning algorithm with different hyperparameter settings (\textit{here:} three) to fit models to the training data.

\paragraph{Step 3.} Next, we evaluate the performance of our models on the validation set. This step illustrates the model selection stage; after comparing the performance estimates, we choose the hyperparameters settings associated with the best performance. Note that we often merge steps two and three in practice: we fit a model and compute its performance before moving on to the next model in order to avoid keeping all fitted models in memory.

\paragraph{Step 4.} As discussed earlier, the performance estimates may suffer from pessimistic bias if the training set is too small. Thus, we can merge the training and validation set after model selection and use the best hyperparameter settings from the previous step to fit a model to this larger dataset.

\paragraph{Step 5.} Now, we can use the independent test set to estimate the generalization performance our model. Remember that the purpose of the test set is to simulate new data that the model has not seen before. Re-using this test set may result in an overoptimistic bias in our estimate of the model's generalization performance.

\paragraph{Step 6.} Finally, we can make use of all our data -- merging training and test set-- and fit a model to all data points for real-world use.

Note that fitting the model on all available data might yield a model that is likely slightly different from the model evaluated in Step 5. However, in theory, using all data (that is, training and test data) to fit the model should only improve its performance. Under this assumption, the evaluated performance from Step 5 might slightly underestimate the performance of the model fitted in Step 6. (If we use test data for fitting, we do not have data left to evaluate the model, unless we collect new data.) In real-world applications, having the "best possible" model is often desired -- or in other words, we do not mind if we slightly underestimated its performance. In any case, we can regard this sixth step as optional.

\subsection{Introduction to $k$-fold Cross-Validation}

\begin{figure}[!htb]
 \centering
    \includegraphics[width=0.85\linewidth]{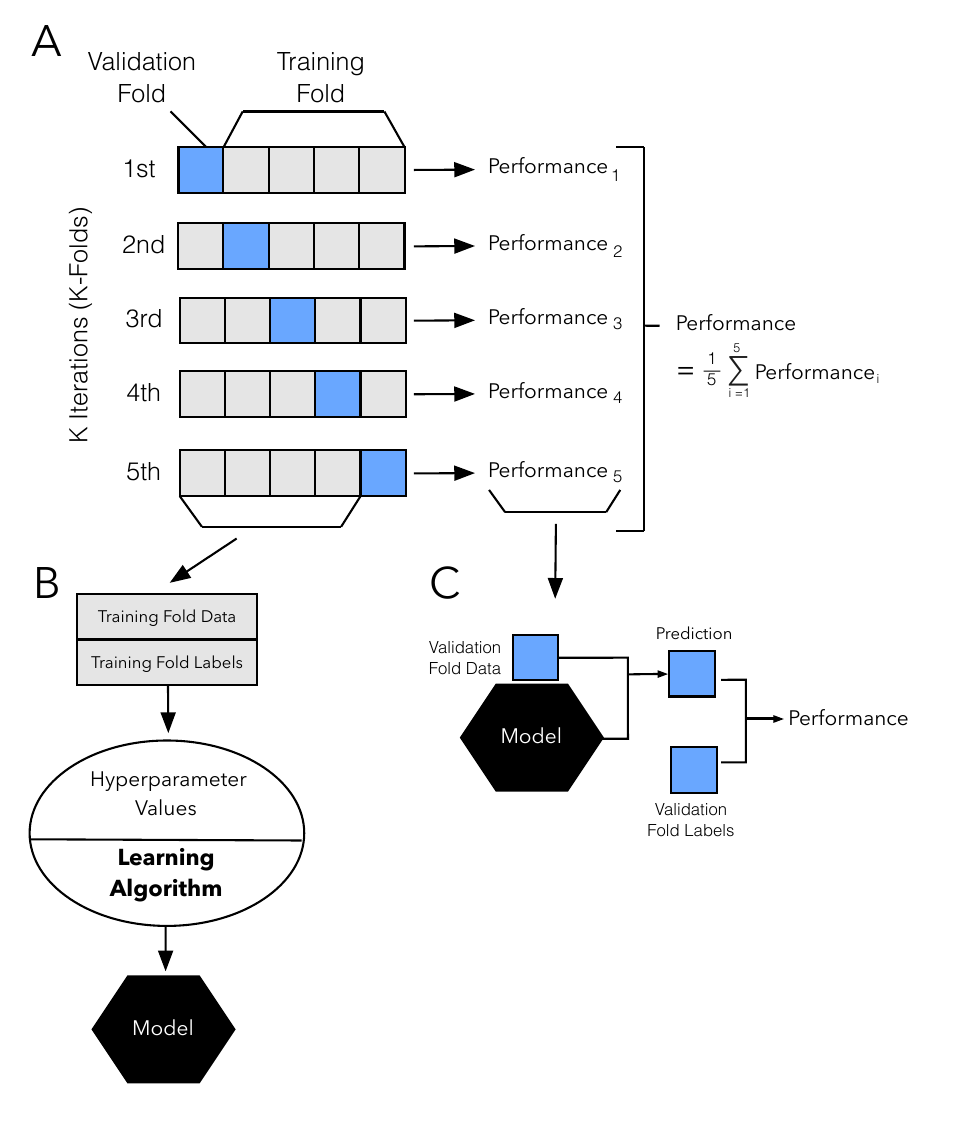}
    \caption{Illustration of the $k$-fold cross-validation procedure.}
    \label{fig13:kfold}
\end{figure}

It is about time to introduce the probably most common technique for model evaluation and model selection in machine learning practice: $k$-fold cross-validation. The term cross-validation is used loosely in literature, where practitioners and researchers sometimes refer to the train/test holdout method as a cross-validation technique. However, it might make more sense to think of cross-validation as a crossing over of training and validation stages in successive rounds. Here, the main idea behind cross-validation is that each sample in our dataset has the opportunity of being tested. $k$-fold cross-validation is a special case of cross-validation where we iterate over a dataset set $k$ times. In each round, we split the dataset into $k$ parts: one part is used for validation, and the remaining $k-1$ parts are merged into a training subset for model evaluation as shown in Figure \ref{fig13:kfold} , which illustrates the process of 5-fold cross-validation.

Just as in the "two-way" holdout method (Section \ref{holdout-validation}), we use a learning algorithm with fixed hyperparameter settings to fit models to the training folds in each iteration when we use the $k$-fold cross-validation method for \textit{model evaluation}. In 5-fold cross-validation, this procedure will result in five different models fitted; these models were fit to distinct yet partly overlapping training sets and evaluated on non-overlapping validation sets. Eventually, we compute the cross-validation performance as the arithmetic mean over the $k$ performance estimates from the validation sets.

We saw the main difference between the "two-way" holdout method and $k$-fold cross validation: $k$-fold cross-validation uses all data for training and testing. The idea behind this approach is to reduce the pessimistic bias by using more training data in contrast to setting aside a relatively large portion of the dataset as test data. And in contrast to the repeated holdout method, which was discussed in Section \ref{sec:boostrapping-and-uncertainties}, test folds in $k$-fold cross-validation are not overlapping. In repeated holdout, the repeated use of samples for testing results in performance estimates that become dependent between rounds; this dependence can be problematic for statistical comparisons, which we will be discussed in Section  \ref{sec:algo-comparison}. Also, $k$-fold cross-validation guarantees that each sample is used for validation in contrast to the repeated holdout-method, where some samples may never be part of the test set.

This section introduced $k$-fold cross-validation for \textit{model evaluation}. In practice, however, $k$-fold cross-validation is more commonly used for model selection or algorithm selection. $k$-fold cross-validation for model selection is a topic that we will be covered in the next sections, and algorithm selection will be discussed in detail throughout Section  \ref{sec:algo-comparison}.

\subsection{Special Cases: 2-Fold and Leave-One-Out Cross-Validation}
\label{sec:loocv}

At this point, you may wonder why $k=5$ was chosen to illustrate $k$-fold cross-validation in the previous section. One reason is that it makes it easier to illustrate $k$-fold cross-validation compactly. Moreover, $k=5$ is also a common choice in practice, since it is computationally less expensive compared to larger values of $k$. If $k$ is too small, though, the pessimistic bias of the performance estimate may increase (since less training data is available for model fitting), and the variance of the estimate may increase as well since the model is more sensitive to how the data was split (in the next sections, experiments will be discussed that suggest $k=10$ as a good choice for $k$). 

\begin{figure}[!htb]
 \centering
    \includegraphics[width=0.65\linewidth]{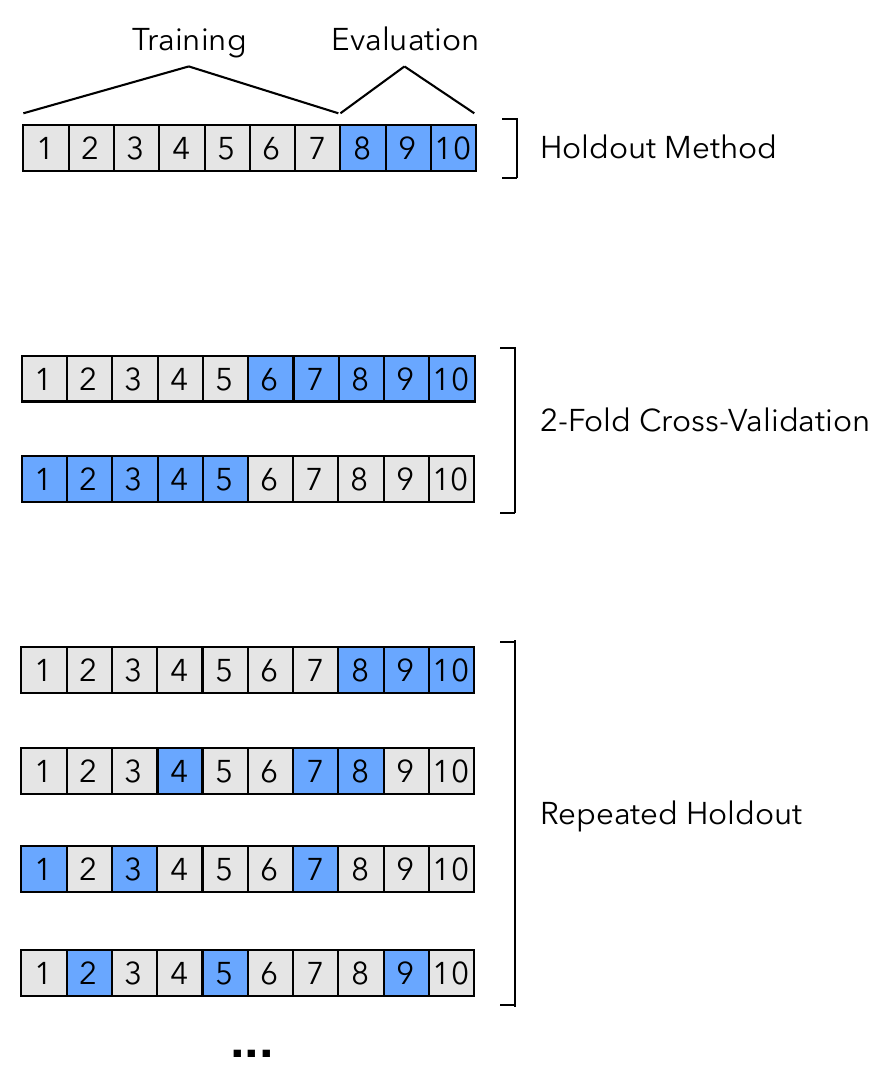}
    \caption{Comparison of the holdout method, 2-fold cross-validation, and the repeated holdout method.}
    \label{fig14:holdout-vs-2fold}
\end{figure}

In fact, there are two prominent, special cases of $k$-fold cross validation: $k=2$ and $k=n$. Most literature describes 2-fold cross-validation as being equal to the holdout method. However, this statement would only be true if we performed the holdout method by rotating the training and validation set in two rounds (for instance, using exactly 50\% data for training and 50\% of the examples for validation in each round, swapping these sets, repeating the training and evaluation procedure, and eventually computing the performance estimate as the arithmetic mean of the two performance estimates on the validation sets). Given how the holdout method is most commonly used though, this article describes the holdout method and 2-fold cross-validation as two different processes as illustrated in Figure \ref{fig14:holdout-vs-2fold}.

\begin{figure}[!htb]
 \centering
    \includegraphics[width=0.45\linewidth]{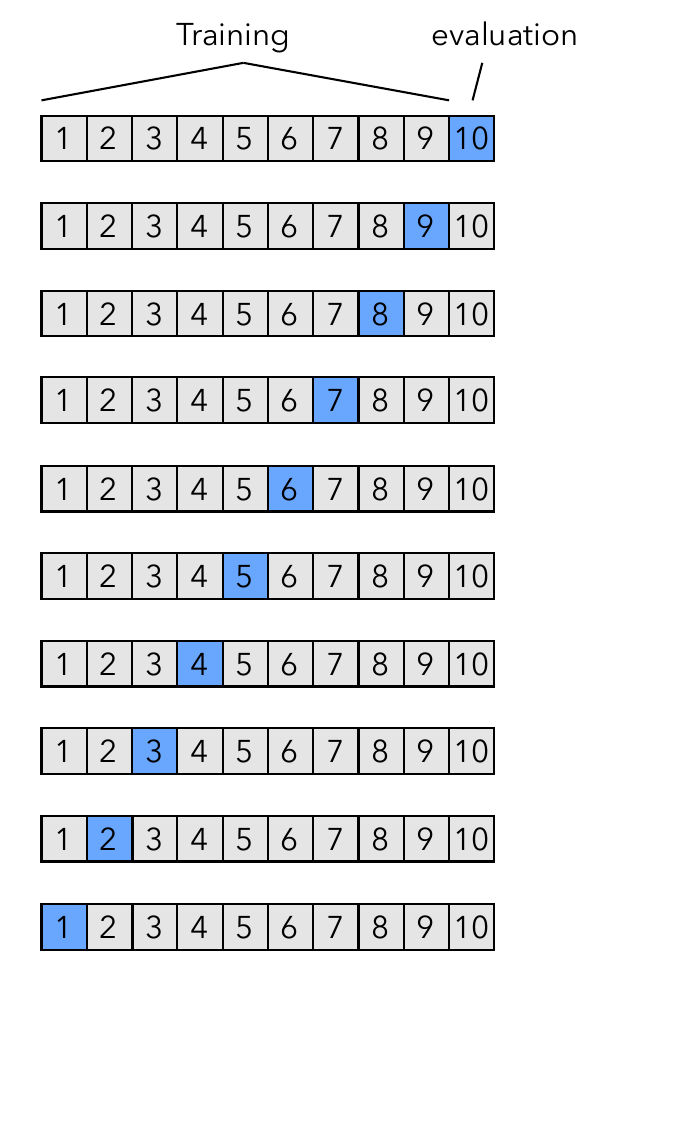}
    \caption{Illustration of leave-one-out cross-validation.}
    \label{fig15:loocv}
\end{figure}

Now, if we set $k=n$, that is, if we set the number of folds as being equal to the number of training instances, we refer to the $k$-fold cross-validation process as \textit{Leave-One-Out Cross-Validation} (LOOCV), which is illustrated in Figure \ref{fig15:loocv}. In each iteration during LOOCV, we fit a model to $n-1$ samples of the dataset and evaluate it on the single, remaining data point. Although this process is computationally expensive, given that we have $n$ iterations, it can be useful for small datasets, cases where withholding data from the training set would be too wasteful.

Several studies compared different values of $k$ in $k$-fold cross-validation, analyzing how the choice of $k$ affects the variance and the bias of the estimate. Unfortunately, there is no \textit{Free Lunch} though as shown by Yohsua Bengio and Yves Grandvalet in "No unbiased estimator of the variance of $k$-fold cross-validation:"

\begin{displayquote}
The main theorem shows that there exists no universal (valid under all distributions) unbiased estimator of the variance of $k$-fold cross-validation.

\cite{bengio2004no}
\end{displayquote}

However, we may still be interested in finding a "sweet spot," a value for $k$ that seems to be a good trade-off between variance and bias in most cases, and the bias-variance trade-off discussion will be continued in the next section. For now, let us conclude this section by looking at an interesting research project where Hawkins and others compared performance estimates via LOOCV to the holdout method and recommend the LOOCV over the latter -- if computationally feasible:

\begin{displayquote}
 
 [...] where available sample sizes are modest, holding back compounds for model testing is ill-advised. This fragmentation of the sample harms the calibration and does not give a trustworthy assessment of fit anyway. It is better to use all data for the calibration step and check the fit by cross-validation, making sure that the cross-validation is carried out correctly. [...] The only motivation to rely on the holdout sample rather than cross-validation would be if there was reason to think the cross-validation not trustworthy -- biased or highly variable. But neither theoretical results nor the empiric results sketched here give any reason to disbelieve the cross-validation results. 

\cite{hawkins2003assessing}
\end{displayquote}

\begin{table}[]
\centering
\caption{Summary of the findings from the LOOCV vs. holdout comparison study conducted by Hawkins \textit{and others}  \cite{hawkins2003assessing}. See text for details.}
\begin{tabular}{lll}
\textbf{Experiment} & \textbf{Mean} & \textbf{Standard deviation} \\
True $R^2$ --- $q^2$   & 0.010         & 0.149                       \\
True $R^2$ --- hold 50 & 0.028         & 0.184                       \\
True $R^2$ --- hold 20 & 0.055         & 0.305                       \\
True $R^2$ --- hold 10 & 0.123         & 0.504                      
\end{tabular}
\label{tab:hawkins-study}
\end{table}

The conclusions in the previous quotation are partly based on the experiments carried out in this study using a 469-sample dataset, and Table \ref{tab:hawkins-study} summarizes the findings in a comparison of different Ridge Regression models evaluated by LOOCV and the holdout method \cite{hawkins2003assessing}. The first row corresponds to an experiment where the researchers used LOOCV to fit regression models to 100-example training subsets. The reported "mean" refers to the averaged difference between the true coefficients of determination ($R^2$) and the coefficients obtained via LOOCV (here called $q^2$) after repeating this procedure on different 100-example training sets and averaging the results. In rows 2-4, the researchers used the holdout method for fitting models to the 100-example training sets, and they evaluated the performances on holdout sets of sizes 10, 20, and 50 samples. Each experiment was repeated 75 times, and the \textit{mean} column shows the average difference between the estimated $R^2$ and the true $R^2$ values. As we can see, the estimates obtained via LOOCV ($q^2$) are the closest to the true $R^2$ on average. The estimates obtained from the 50-example test set via the holdout method are also passable, though. Based on these particular experiments, we may agree with the researchers' conclusion:

\begin{displayquote}
Taking the third of these points, if you have 150 or more compounds available, then you can certainly make a random split into 100 for calibration and 50 or more for testing. However it is hard to see why you would want to do this.

\cite{hawkins2003assessing}
\end{displayquote}

One reason why we may prefer the holdout method may be concerns about computational efficiency, if the dataset is sufficiently large. As a rule of thumb, we can say that the pessimistic bias and large variance concerns are less problematic the larger the dataset. Moreover, it is not uncommon to repeat the $k$-fold cross-validation procedure with different random seeds in hope to obtain a "more robust" estimate. For instance, if we repeated a 5-fold cross-validation run 100 times, we would compute the performance estimate for 500 test folds report the cross-validation performance as the arithmetic mean of these 500 folds. (Although this is commonly done in practice, we note that the test folds are now overlapping.) However, there is no point in repeating LOOCV, since LOOCV always produces the same splits.

\subsection{$k$-fold Cross-Validation and the Bias-Variance Trade-off}

Based on the study by Hawkins \textit{and others} \cite{hawkins2003assessing} discussed in Section \ref{sec:loocv} we may prefer LOOCV over single train/test splits via the holdout method for small and moderately sized datasets. In addition, we can think of the LOOCV estimate as being approximately unbiased: the pessimistic bias of LOOCV ($k=n$) is intuitively lower compared $k<n$-fold cross-validation, since almost all (for instance, $n-1$) training samples are available for model fitting.

While LOOCV is almost unbiased, one downside of using LOOCV over $k$-fold cross-validation with $k<n$ is the large variance of the LOOCV estimate. First, we have to note that LOOCV is \textit{defect} when using a discontinuous loss-function such as the 0-1 loss in classification or even in continuous loss functions such as the mean-squared-error. It is said that LOOCV "[LOOCV has] high variance because the test set only contains one sample"  \cite{tan2005datamining} and "[LOOCV] is highly variable, since it is based upon a single observation (x1, y1)" \cite{gareth2013stat}. These statements are certainly true if we refer to the variance between folds. Remember that if we use the 0-1 loss function (the prediction is either correct or not), we could consider each prediction as a Bernoulli trial, and the number of correct predictions $X$ if following a binomial distribution $X \approx B(n, p)$, where $n \in \mathbb{N}$ and $ p \in [0,1]$; the variance of a binomial distribution is defined as 

\begin{equation}
\sigma^2 = np(1-p)
\end{equation}

We can estimate the variability of a statistic (\textit{here}: the performance of the model) from the variability of that statistic between subsamples. Obviously though, the variance between folds is a poor estimate of the variance of the LOOCV estimate -- the variability due to randomness in our training data. Now, when we are talking about the variance of LOOCV, we typically mean the difference in the results that we would get if we repeated the resampling procedure multiple times on different data samples from the underlying distribution. In this context interesting point has been made by Hastie, Tibshirani, and Friedman:

\begin{displayquote}
With $k=n$, the cross-validation estimator is approximately unbiased for the true (expected) prediction error, but can have high variance because the $n$ "training sets" are so similar to one another. 

\cite{hastie2009}
\end{displayquote}

Or in other words, we can attribute the high variance to the well-known fact that the mean of highly correlated variables has a higher variance than the mean of variables that are not highly correlated. Maybe, this can intuitively be explained by looking at the relationship between covariance ($\text{cov}$) and variance ($\sigma^2$):

\begin{equation}
\text{cov}_{X, X} = \sigma^2_{X}.
\label{cov-eq-var}
\end{equation}

or

\begin{equation}
\text{cov}_{X, X} = E\left[(X - \mu)^2\right] = \sigma^{2}_{X}
\end{equation}

if we let $\mu = E(X)$. 

And the relationship between covariance $\text{cov}_{X, Y}$ and correlation $\rho_{X, Y}$ (X and Y are random variables)  is defined as 

\begin{equation}
\text{cov}_{X, Y} = \rho_{X, Y} \; \sigma_X \sigma_Y,
\end{equation}

where

\begin{equation}
\text{cov}_{X, Y} = E [(X - \mu_X)(Y - \mu_Y)]
\end{equation}

and

\begin{equation}
\rho_{X, Y} = E [(X - \mu_X)(Y - \mu_Y)] / (\sigma_X \sigma_Y).
\end{equation}

The large variance that is often associated with LOOCV has also been observed in empirical studies \cite{kohavi1995}.

Now that we established that LOOCV estimates are generally associated with a large variance and a small bias, how does this method compare to $k$-fold cross-validation with other choices for $k$ and the bootstrap method? Section \ref{sec:boostrapping-and-uncertainties} discussed the pessimistic bias of the standard bootstrap method, where the training set asymptotically (only) contains 0.632 of the samples from the original dataset; 2- or 3-fold cross-validation has about the same problem (the .632 bootstrap that was designed to address this pessimistic bias issue). However, Kohavi also observed in his experiments \cite{kohavi1995} that the bias in bootstrap was still extremely large for certain real-world datasets (now, optimistically biased) compared to $k$-fold cross-validation. Eventually, Kohavi's experiments on various real-world datasets suggest that 10-fold cross-validation offers the best trade-off between bias and variance. Furthermore, other researchers found that repeating $k$-fold cross-validation can increase the precision of the estimates while still maintaining a small bias \cite{molinaro2005prediction,kim2009estimating}.

Before moving on to model selection in the next section, the following points shall summarize the discussion of the bias-variance trade-off, by listing the general trends when increasing the number of folds or $k$:

\begin{itemize}
\item The bias of the performance estimator decreases (more accurate)
\item The variance of the performance estimators increases (more variability)
\item The computational cost increases (more iterations, larger training sets during fitting)
\item Exception: decreasing the value of $k$ in $k$-fold cross-validation to small values (for example, 2 or 3) also increases the variance on small datasets due to random sampling effects.
\end{itemize}

\subsection{Model Selection via $k$-fold Cross-Validation}

\begin{figure}[!htb]
 \centering
    \includegraphics[width=0.85\linewidth]{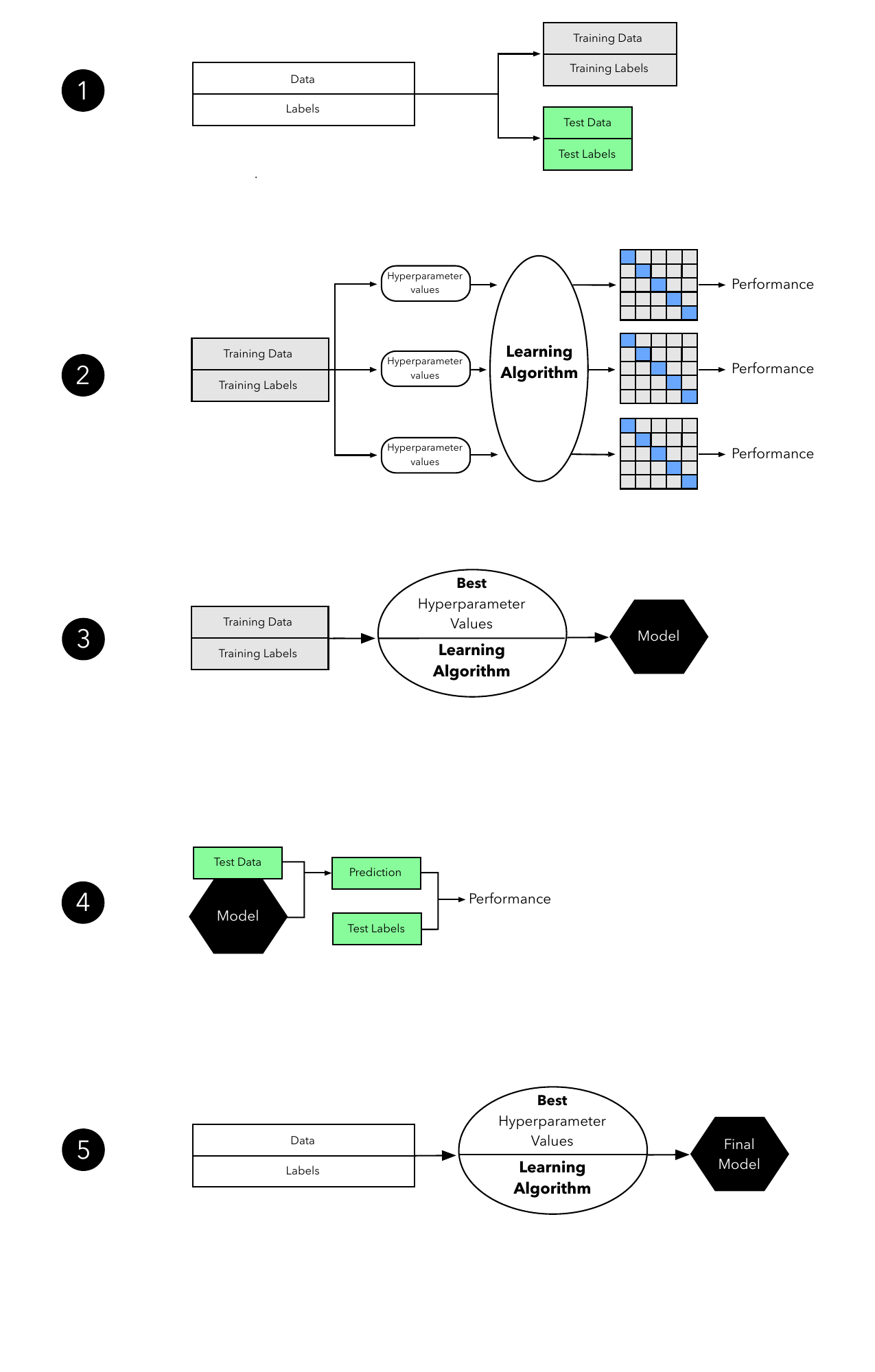}
    \caption{Illustration of $k$-fold cross-validation for model selection.}
    \label{fig15:kfold-selection}
\end{figure}

Previous sections introduced $k$-fold cross-validation for model \textit{evaluation}. Now, this section discusses how to use the $k$-fold cross-validation method for model \textit{selection}. Again, the key idea is to keep an independent test dataset, that we withhold from during training and model selection, to avoid the leaking of test data in the training stage. This procedure is outlined in Figure \ref{fig15:kfold-selection}.

Although Figure \ref{fig15:kfold-selection} might seem complicated at first, the process is quite simple and analogous to the "three-way holdout" workflow that we discussed at the beginning of this article. The following paragraphs will discuss Figure \ref{fig15:kfold-selection} step by step.

\paragraph{Step 1.} Similar to the holdout method, we split the dataset into two parts, a training and an independent test set; we tuck away the test set for the final model evaluation step at the end (Step 4).

\paragraph{Step 2.} In this second step, we can now experiment with various hyperparameter settings; we could use Bayesian optimization, randomized search, or grid search, for example. For each hyperparameter configuration, we apply the $k$-fold cross-validation method on the training set, resulting in multiple models and performance estimates.

\paragraph{Step 3.} Taking the hyperparameter settings that produced the best results in the $k$-fold cross-validation procedure, we can then use the complete training set for model fitting with these settings.

\paragraph{Step 4.} Now, we use the independent test set that we withheld earlier (Step 1); we use this test set to evaluate the model that we obtained from Step 3.

\paragraph{Step 5.} Finally, after we completed the evaluation stage, we can optionally fit a model to all data (training and test datasets combined), which could be the model for (the so-called) deployment.

\subsection{A Note About Model Selection and Large Datasets}

When we browse the deep learning literature, we often find that that the 3-way holdout method is the method of choice when it comes to model evaluation; it is also common in older (non-deep learning literature) as well. As mentioned earlier, the three-way holdout may be preferred over $k$-fold cross-validation since the former is computationally cheap in comparison. Aside from computational efficiency concerns, we only use deep learning algorithms when we have relatively large sample sizes anyway, scenarios where we do not have to worry about high variance -- the sensitivity of our estimates towards how we split the dataset for training, validation, and testing -- so much. Thus, it is fine to use the holdout method with a training, validation, and test split over the $k$-fold cross-validation for model selection if the dataset is relatively large.

\subsection{A Note About Feature Selection During Model Selection}

Note that if we normalize data or select features, we typically perform these operations inside the $k$-fold cross-validation loop in contrast to applying these steps to the whole dataset upfront before splitting the data into folds. Feature selection inside the cross-validation loop reduces the bias through overfitting, since it avoids peaking at the test data information during the training stage. However, feature selection inside the cross-validation loop may lead to an overly pessimistic estimate, since less data is available for training. A more detailed discussion on this topic, whether to perform feature selection inside or outside the cross-validation loop, can be found in Refaeilzadeh's "On comparison of feature selection algorithms" \cite{refaeilzadeh2007comparison}.

\subsection{The Law of Parsimony}

Now that we discussed model selection in the previous section, let us take a moment and consider the Law of Parsimony, which is also known as Occam's Razor: "Among competing hypotheses, the one with the fewest assumptions should be selected." In model selection practice, Occam's razor can be applied, for example, by using the \textit{one-standard error method} \cite{breiman1984classification} as follows:

\begin{enumerate}
  \item Consider the numerically optimal estimate and its standard error.
  \item Select the model whose performance is within one standard error of the value obtained in step 1.
\end{enumerate}

Although, we may prefer simpler models for several reasons, Pedro Domingos made a good point regarding the performance of "complex" models. Here is an excerpt from his article, "Ten Myths About Machine Learning\footnote{https://medium.com/@pedromdd/ten-myths-about-machine-learning-d888b48334a3}:"

\begin{displayquote}
Simpler models are more accurate. This belief is sometimes equated with Occam's razor, but the razor only says that simpler explanations are preferable, not why. They're preferable because they're easier to understand, remember, and reason with. Sometimes the simplest hypothesis consistent with the data is less accurate for prediction than a more complicated one. Some of the most powerful learning algorithms output models that seem gratuitously elaborate? -- sometimes even continuing to add to them after they've perfectly fit the data -- but that's how they beat the less powerful ones.
\end{displayquote}

Again, there are several reasons why we may prefer a simpler model if its performance is within a certain, acceptable range -- for example, using the one-standard error method. Although a simpler model may not be the most "accurate" one, it may be computationally more efficient, easier to implement, and easier to understand and reason with compared to more complicated alternatives.

To see how the one-standard error method works in practice, let us consider a simple toy dataset: 300 training examples, concentric circles, and a uniform class distribution (150 samples from class 1 and 150 samples from class 2).

\begin{figure}[!htb]
 \centering
    \includegraphics[width=0.85\linewidth]{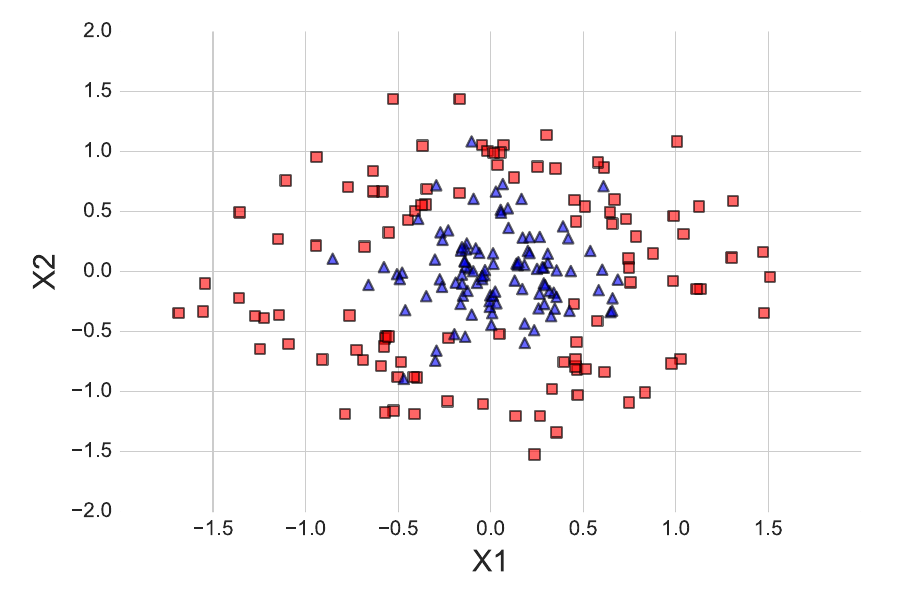}
    \caption{Concentric circles dataset with 210 training examples and uniform class distribution.}
    \label{fig16:concentric-1}
\end{figure}

The concentric circles dataset is then split into two parts, 70\% training data and 30\% test data, using stratification to maintain equal class proportions. The 210 samples from the training dataset are shown in Figure \ref{fig16:concentric-1}. 

Now, let us assume that our goal is to optimize the $\gamma$ (gamma) hyperparameter of a Support Vector Machine (SVM) with a non-linear Radial Basis Function-kernel (RBF-kernel), where $\gamma$ is the free parameter of the Gaussian RBF:

\begin{equation}
K(x_i, x_j) = exp(-\gamma || x_i - x_j ||^2), \gamma > 0.
\end{equation}

(Intuitively, we can think of  $\gamma$ as a parameter that controls the influence of single training samples on the decision boundary.)

\begin{figure}[!htb]
 \centering
    \includegraphics[width=1.\linewidth]{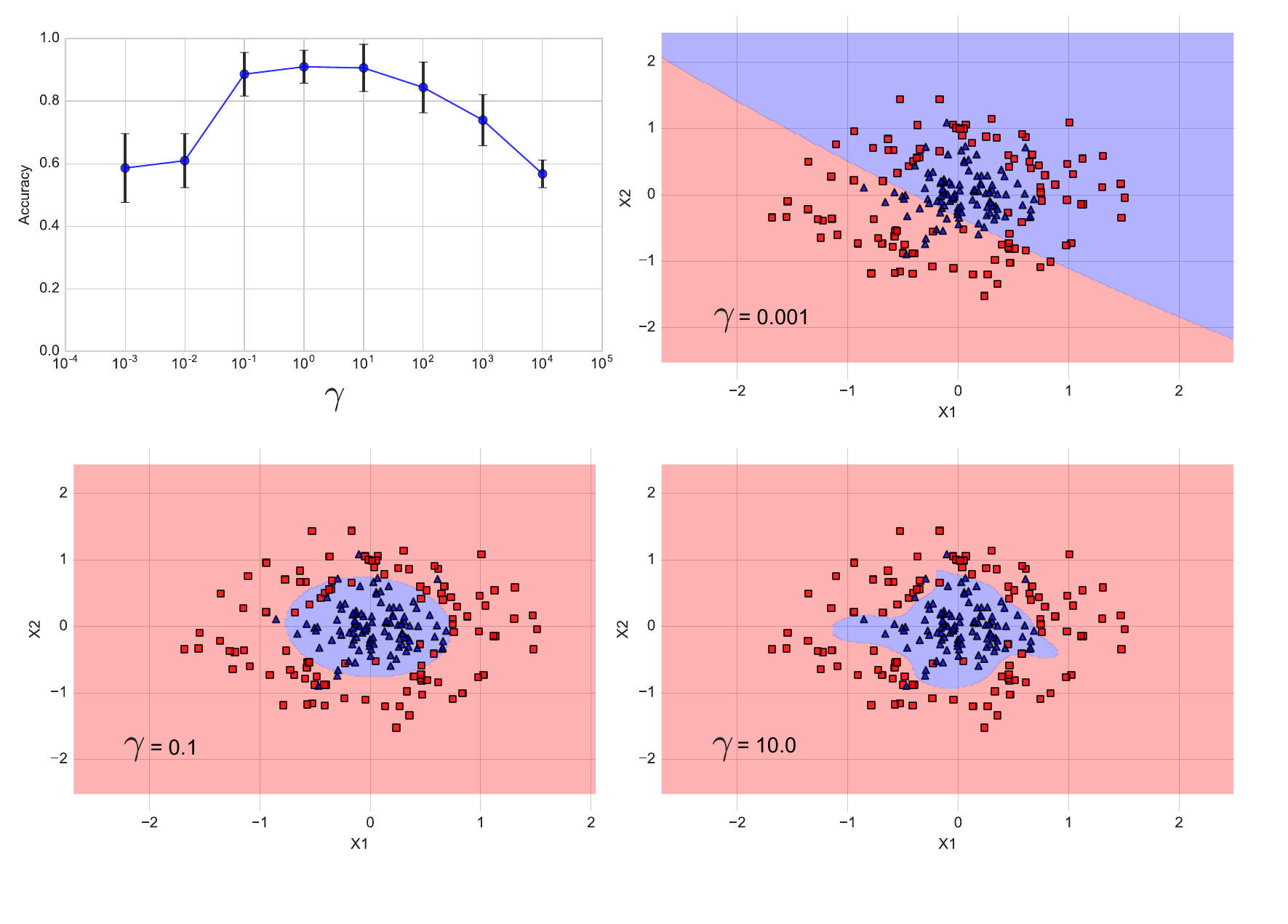}
    \caption{Performance estimates and decision regions of an RBF-kernel SVM with stratified 10-fold cross-validation for different $\gamma$ values. Error bars represent the standard error of the cross-validation estimates.}
    \label{fig17:concentric-2}
\end{figure}

After running the RBF-kernel SVM algorithm with different $\gamma$ values over the training set, using stratified 10-fold cross-validation, the performance estimates shown in Figure \ref{fig17:concentric-2} were obtained, where the error bars are the standard errors of the cross-validation estimates.

As shown in Figure \ref{fig17:concentric-2}, Choosing $\gamma$ values between 0.1 and 100 resulted in a prediction accuracy of 80\% or more. Furthermore, we can see that $\gamma=10.0$ resulted in a fairly complex decision boundary, and $\gamma=0.001$ resulted in a decision boundary that is too simple to separate the two classes in the concentric circles dataset. In fact, $\gamma=0.1$ seems like a good trade-off between the two aforementioned models ($\gamma=10.0$ and $\gamma=0.1$)  -- the performance of the corresponding model falls within one standard error of the best performing model with $\gamma=0$ or $\gamma=10$.

\subsection{Summary}

There are many ways for evaluating the generalization performance of predictive models. So far, this article covered the holdout method, different flavors of the bootstrap approach, and $k$-fold cross-validation. Using holdout method is absolutely fine for model evaluation when working with relatively large sample sizes. For hyperparameter optimization, we may prefer 10-fold cross-validation, and Leave-One-Out cross-validation is a good option when working with small sample sizes. When it comes to model selection, again, the "three-way" holdout method may be a good choice if the dataset is large, if computational efficiency is a concern; a good alternative is using $k$-fold cross-validation with an independent test set. While the previous sections focused on model evaluation and hyperparameter optimization, Section \ref{sec:algo-comparison} introduces different techniques for comparing different learning algorithms.

The next section will discuss different statistical methods for comparing the performance of different models as well as empirical approaches for comparing different machine learning algorithms.

\section{Algorithm Comparison}
\label{sec:algo-comparison}

\subsection{Overview}

This final section in this article presents overviews of several statistical hypothesis testing approaches, with applications to machine learning model and algorithm comparisons. This includes statistical tests based on target predictions for independent test sets (the downsides of using a single test set for model comparisons was discussed in previous sections) as well as methods for algorithm comparisons by fitting and evaluating models via cross-validation. Lastly, this final section will introduce \textit{nested cross-validation}, which has become a common and recommended a method of choice for algorithm comparisons for small to moderately-sized datasets.

Then, at the end of this article, I provide a list of my personal suggestions concerning model evaluation, selection, and algorithm selection summarizing the several techniques covered in this series of articles.

\subsection{Testing the Difference of Proportions}

There are several different statistical hypothesis testing frameworks that are being used in practice to compare the performance of classification models, including conventional methods such as difference of two proportions (here, the proportions are the estimated generalization accuracies from a test set), for which we can construct 95\% confidence intervals based on the concept of the Normal Approximation to the Binomial that was covered in Section \ref{sec:introduction}. 

Performing a z-score test for two population proportions is inarguably the most straight-forward way to compare to models (but certainly not the best!): In a nutshell, if the 95\% confidence intervals of the accuracies of two models do not overlap, we can reject the null hypothesis that the performance of both classifiers is equal at a confidence level of $\alpha=0.05$ (or 5\% probability). Violations of assumptions aside (for instance that the test set samples are not independent), as Thomas Dietterich noted based on empirical results in a simulated study \cite{dietterich1998approximate}, this test tends to have a high false positive rate (\textit{here:} incorrectly detecting  difference when there is none), which is among the reasons why it is not recommended in practice.

Nonetheless, for the sake of completeness, and since it a commonly used method in practice, the general procedure is outlined below as follows (which also generally applies to the different hypothesis tests presented later): 

\begin{enumerate}
\item formulate the hypothesis to be tested (for instance, the null hypothesis stating that the proportions are the same; consequently, the alternative hypothesis that the proportions are different, if we use a two-tailed test);
\item decide upon a significance threshold (for instance, if the probability of observing a difference more extreme than the one seen is more than 5\%, then we plan to reject the null hypothesis); 
\item analyze the data, compute the test statistic (\textit{here}: $z$-score), and compare its associated $p$-value (probability) to the previously determined significance threshold;
\item based on the $p$-value and significance threshold, either accept or reject the null hypothesis at the given confidence level and interpret the results.
\end{enumerate}

The z-score is computed as the observed difference divided by the square root for their combined variances
$$
z = \frac{ACC_1 - ACC_2}{\sqrt{\sigma_{1}^2 + \sigma_{2}^2}},
$$
where $ACC_1$ is the accuracy of one model and $ACC_2$ is the accuracy of a second model estimated from the test set. Recall that we computed the variance of the estimated accuracy as 
$$
\sigma^2 = \frac{ACC(1-ACC)}{n}
$$
in Section \ref{sec:introduction} and then computed the confidence interval (Normal Approximation Interval) as 
$$
ACC \pm z \times \sigma,
$$

where $z=1.96$ for a 95\% confidence interval.  Comparing the confidence intervals of two accuracy estimates and checking whether they overlap is then analogous to computing the $z$ value for the difference in proportions and comparing the probability ($p$-value) to the chosen significance threshold. So, to compute the z-score directly for the difference of two proportions, $ACC_1$ and $ACC_2$, we pool these proportions (assuming that $ACC_1$ and $ACC_2$ are the performances of two models estimated on two indendent test sets of size $n_1$ and $n_2$, respectively),

$$
ACC_{1, 2} = \frac{ACC_1 \times n_1 + ACC_2 \times n_2}{n_1 + n_2},
$$

and compute the standard deviation as

$$
\sigma_{1,2} = \sqrt{ ACC_{1,2} (1-ACC_{1,2}) \times \left(\frac{1}{n_1} + \frac{1}{n_2}\right)  },
$$

such that we can compute the z-score, 

$$
z = \frac{ACC_1 - ACC_2}{\sigma_{1,2}}.
$$

Since, due to using the same test set (and violating the independence assumption) we have $n_1 = n_2 = n$, so that we can simplify the z-score computation to 
$$
z =  \frac{ACC_1 - ACC_2}{\sqrt{2\sigma^2}} = \frac{ACC_1 - ACC_2}{\sqrt{2\cdot ACC_{1,2}(1-ACC_{1,2}))/n}}.
$$
where $ACC_{1, 2}$ is simply $(ACC_1 + ACC_2)/2$.

In the second step, based on the computed $z$ value (this assumes the test errors are independent, which is usually violated in practice as we use the same test set) we can reject the null hypothesis that the pair of models has equal performance (here, measured in "classification accuracy") at an $\alpha=0.05$ level if $|z|$ is higher than 1.96. Alternatively, if we want to put in the extra work, we can compute the area under the standard normal cumulative distribution at the z-score threshold. If we find this $p$-value is smaller than a significance level we set before conducting the test, then we can reject the null hypothesis at that given significance level.

The problem with this test though is that we use the same test set to compute the accuracy of the two classifiers; thus, it might be better to use a paired test such as a paired sample $t$-test, but a more robust alternative is the McNemar test illustrated in the next section.

\subsection{Comparing Two Models with the McNemar Test}

So, instead of using the "difference of proportions" test, Dietterich \cite{dietterich1998approximate} found that the McNemar test is to be preferred. The McNemar test, introduced by Quinn McNemar in 1947 \cite{mcnemar1947note}, is a non-parametric statistical test for paired comparisons that can be applied to compare the performance of two machine learning classifiers. 

\begin{figure}[htb!]
 \centering
    \includegraphics[width=0.5\linewidth]{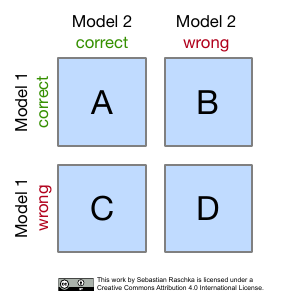}
    \caption{Confusion matrix layout in context of McNemar's test.}
    \label{fig18:mcnemar-layout}
\end{figure}

Often, McNemar's test is also referred to as "within-subjects chi-squared test," and it is applied to paired nominal data based on a version of 2x2 confusion matrix (sometimes also referred to as \textit{2x2 contingency table}) that compares the predictions of two models to each other (not be confused with the typical confusion matrices encountered in machine learning, which are listing false positive, true positive, false negative, and true negative counts of a single model). The layout of the 2x2 confusion matrix suitable for McNemar's test is shown in Figure \ref{fig18:mcnemar-layout}.

Given such a 2x2 confusion matrix as shown in Figure \ref{fig18:mcnemar-layout}, we can compute the accuracy of a \textit{Model 1} via $(A+B) / (A+B+C+D)$, where $A+B+C+D$ is the total number of test examples $n$. Similarly, we can compute the accuracy of Model 2 as $(A+C) / n$. The most interesting numbers in this table are in cells B and C, though, as A and D merely count the number of samples where both \textit{Model 1} and \textit{Model 2} made correct or wrong predictions, respectively. Cells B and C (the off-diagonal entries), however, tell us how the models differ. To illustrate this point, let us take a look at the following example:

\begin{figure}[htb!]
 \centering
    \includegraphics[width=\linewidth]{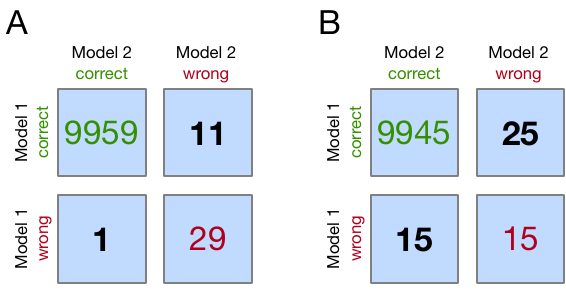}
    \caption{Confusion matrix for exemplary classification outcomes of two models, \textit{Model 1} and \textit{Model 2}.}
    \label{fig19:mcnemar-example}
\end{figure}

In both subpanels, A and B, in Figure \ref{fig19:mcnemar-example}, the accuracy of \textit{Model 1} and \textit{Model 2} are 99.6\% and 99.7\%, respectively. 

\begin{itemize}
  \item Model 1 accuracy subpanel A: $(9959+11) / 10000 \times 100\% = 99.7\%$
  \item Model 1 accuracy subpanel B: $(9945+25) / 10000 \times 100\% = 99.7\%$
  \item Model 2 accuracy subpanel A: $(9959+1) / 10000 \times 100\% = 99.6\%$
  \item Model 2 accuracy subpanel B: $(9945+15) / 10000 \times 100\% = 99.6\%$
\end{itemize}

Now, in subpanel A, we can see that Model 1 got 11 predictions right that Model 2 got wrong. Vice versa, Model 2 got one prediction right that Model 1 got wrong. Thus, based on this 11:1 ratio, we may conclude, based on our intuition, that Model 1 performs substantially better than Model 2. However, in subpanel B, the Model 1:Model 2 ratio is 25:15, which is less conclusive about which model is the better one to choose.  This is a good example where McNemar's test can come in handy.

In McNemar's Test, we formulate the null hypothesis that the probabilities $p(B)$ and $p(C)$ -- where $B$ and $C$ refer to the confusion matrix cells introduced in an earlier figure -- are the same, or in simplified terms: None of the two models performs better than the other. Thus, we might consider the alternative hypothesis that the performances of the two models are not equal. 

The McNemar test statistic ("chi-squared") can be computed as follows:

$$\chi^2 = \frac{(B-C)^2}{B+C}.$$

After setting a significance threshold, for example, $\alpha=0.05$, we can compute a $p$-value -- assuming that the null hypothesis is true, the $p$-value is the probability of observing the given empirical (or a larger) $\chi^2$-squared value. If the $p$-value is lower than our chosen significance level, we can reject the null hypothesis that the two models' performances are equal.

Since the McNemar test statistic, $\chi^2$, follows a $\chi^2$ distribution with one degree of freedom (assuming the null hypothesis and relatively large numbers in cells B and C, say > 25), we can now use our favorite software package to "look up" the (1-tail) probability via the $\chi^2$ probability distribution with one degree of freedom.

If we did this for scenario B in the previous figure ($\chi^2=2.5$), we would obtain a $p$-value of 0.1138, which is larger than our significance threshold, and thus, we cannot reject the null hypothesis. Now, if we computed the $p$-value for scenario A ($\chi^2=8.3$), we would obtain a $p$-value of 0.0039, which is below the set significance threshold ($\alpha=0.05$) and leads to the rejection of the null hypothesis; we can conclude that the models' performances are different (for instance, Model 1 performs better than Model 2).

Approximately one year after Quinn McNemar published the McNemar Test \cite{mcnemar1947note}, Allen L. Edwards \cite{edwards1948note} proposed a continuity corrected version, which is the more commonly used variant today:

$$
\chi^2 = \frac{\big(| B - C| -1 \big)^2}{B+C}.
$$

In particular, Edwards wrote:

\begin{displayquote}
This correction will have the apparent result of reducing the absolute value of the difference, [B - C], by unity.
\end{displayquote}

According to Edward, this continuity correction increases the usefulness and accuracy of McNemar's test if we are dealing with discrete frequencies and the data is evaluated regarding the chi-squared distribution.

A function for using McNemar's test is implemented in MLxtend \cite{raschka2018mlxtend}.\footnote{\url{http://rasbt.github.io/mlxtend/user_guide/evaluate/mcnemar/}}

\subsection{Exact $p$-Values via the Binomial Test}

While McNemar's test approximates the $p$-values reasonably well if the values in cells B and C are larger than 50 (referring to the 2x2 confusion matrix shown earlier), for example, it makes sense to use a computationally more expensive binomial test to compute the exact $p$-values if the values of B and C  are relatively small -- since the chi-squared value from McNemar's test may not be well-approximated by the chi-squared distribution. 

The exact $p$-value can be computed as follows (based on the fact that McNemar's test, under the null hypothesis, is essentially a binomial test with proportion 0.5):

$$p = 2 \sum^{n}_{i=B} \binom{n}{i} 0.5^i (1 - 0.5)^{n-i},$$

where $n=B+C$, and the factor 2 is used to compute the two-sided $p$-value (here, $n$ is not to be confused with the test set size $n$).

\begin{figure}[htb!]
 \centering
    \includegraphics[width=\linewidth]{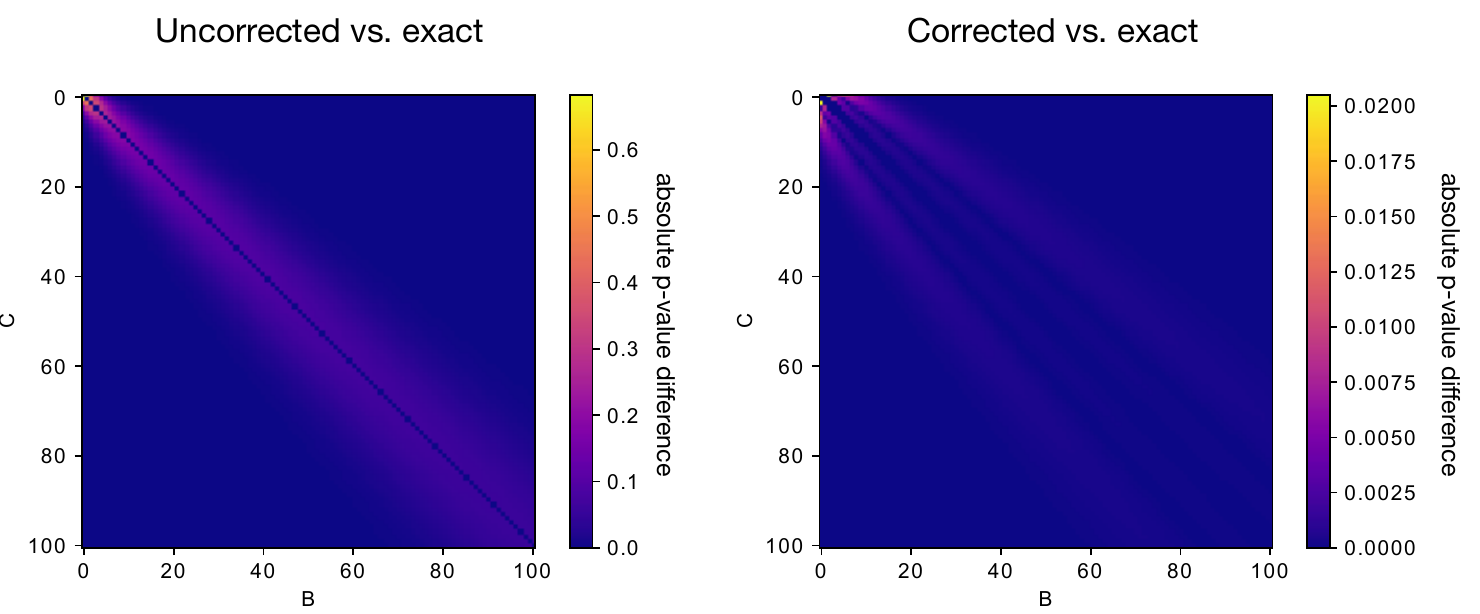}
    \caption{Differences between the McNemar approximation of the chi-squared value (with and without Edward's continuity correction) and the exact $p$-values computed via the binomial test.}
    \label{fig20:pvalue-diff}
\end{figure}

The heat map shown in Figure \ref{fig20:pvalue-diff} illustrates the differences between the McNemar approximation of the chi-squared value (with and without Edward's continuity correction) to the exact $p$-values computed via the binomial test.

As we can see in Figure \ref{fig20:pvalue-diff}, the $p$-values from the continuity-corrected version of McNemar's test are almost identical to the $p$-values from a binomial test if both B and C are larger than 50. (The MLxtend function, "mcnemar," provides different options for toggling between the regular and the corrected McNemar test as well as including an option to compute the exact $p$-values.\footnote{\url{http://rasbt.github.io/mlxtend/user_guide/evaluate/mcnemar/}})

\subsection{Multiple Hypotheses Testing}

In the previous section, we discussed how we could compare two machine learning classifiers using McNemar's test. However, in practice, we often have more than two models that we like to compare based on their estimated generalization performance -- for instance, the predictions on an independent test set. Now, applying the testing procedure described earlier multiple times will result in a typical issue called "multiple hypotheses testing." A common approach for dealing with such scenarios is the following:

\begin{enumerate}
  \item Conduct an omnibus test under the null hypothesis that there is no difference between the classification accuracies.
  \item If the omnibus test led to the rejection of the null hypothesis, conduct pairwise post hoc tests, with adjustments for multiple comparisons, to determine where the differences between the model performances occurred. (Here, we could use McNemar's test, for example.)
\end{enumerate}

Omnibus tests are statistical tests designed to check whether random samples depart from a null hypothesis. A popular example of an omnibus test is the so-called Analysis of Variance (ANOVA), which is a procedure for analyzing the differences between group means. In other words, ANOVA is commonly used for testing the significance of the null hypothesis that the means of several groups are equal. To compare multiple machine learning models, Cochran's $Q$ test would be a possible choice, which is essentially a generalized version of McNemar's test for three or more models. However, omnibus tests are overall significance tests that do not provide information about how the different models differ -- omnibus tests such as Cochran's $Q$ only tell us that a group of models differs or not.

Since omnibus tests can tell us \textit{that} but not \textit{how} models differ, we can perform post hoc tests if an omnibus test leads to the rejection of the null hypothesis. In other words, if we successfully rejected the null hypothesis that the performance of three or models is the same at a predetermined significance threshold, we may conclude that there is at least one significant difference among the different models.

By definition, \textit{post hoc} testing procedures do not require any prior plan for testing. Hence, \textit{post hoc} testing procedures may have a bad reputation and may be regarded as "fishing expeditions" or "searching for the needle in the haystack," because no hypothesis is clear beforehand in terms of which models should be compared, so that it is required to compare all possible pairs of models with each other, which leads to the multiple hypothesis problems that we briefly discussed in Section \ref{sec:cross-val}. \textbf{However, please keep in mind that these are all approximations and everything concerning statistical tests and reusing test sets (independence violation) should be taken with (at least) a grain of salt.}

For the post hoc procedure, we could use one of the many correction terms, for example, Bonferroni's correction \cite{bonferroni1936teoria,dunn1961multiple} for multiple hypothesis testing. In a nutshell, where a given significance threshold (or $\alpha$-level) may be appropriate for a single comparison between to models, it is not suitable for multiple pair-wise comparisons. For instance, using the Bonferroni correction is a means to reduce the false positive rate in multiple comparison tests by adjusting the significance threshold to be more conservative.

The next two sections will discuss two omnibus tests, Cochran's $Q$ test and the $F$-test for comparing multiple classifiers on the same test set proposed by Looney \cite{looney1988statistical}.

\subsection{Cochran's $Q$ Test for Comparing the Performance of Multiple Classifiers}

Cochran's $Q$ test can be regarded as a generalized version of McNemar's test that can be applied to compare three or more classifiers. In a sense, Cochran's $Q$ test is similar to ANOVA but for paired nominal data. Similar to ANOVA, it does not provide us with information about which groups (or models) differ -- only tells us that there is a difference among the models.

The test statistic $Q$ is approximately, (similar to McNemar's test), distributed as chi-squared with $M-1$ degrees of freedom, where $M$ is the number of models we evaluate (since $M=2$ for McNemar's test, McNemar's test statistic approximates a chi-squared distribution with one degree of freedom). 

More formally, Cochran's $Q$ test tests the null hypothesis ($H_0$) that there is no difference between the classification accuracies \cite{fleiss2013statistical}: 

$$H_0: ACC_1 = ACC_2 = \ldots = ACC_M.$$

Let $\{C_1, \dots , C_M\}$ be a set of classifiers who have all been tested on the same dataset. If the $M$ classifiers do not differ in terms of their performance, then the following $Q$ statistic is distributed approximately as
"chi-squared" with $M-1$ degrees of freedom:
$$
Q = (M-1) \frac{M \sum^{M}_{i=1}G_{i}^{2} - T^2}{MT - \sum^{n}_{j=1}M_j^2} .
$$
Here, $G_i$ is the number of objects out of the $n$ test examples correctly classified by $C_i= 1, \dots  M$; $M_j$ is the number of classifiers out of $M$ that correctly classified the $j$th example in the test dataset; and $T$ is the total number of correct number of votes among the $M$ classifiers \cite{kuncheva2004combining}:

$$ T = \sum_{i=1}^{M}; \quad G_i = \sum^{N}_{j=1} M_j.$$

To perform Cochran's $Q$ test, we typically organize the classifier predictions in a binary $n \times M$ matrix (number of test examples vs. the number of classifiers). The $ij\text{th}$ entry of such matrix is 0 if a classifier $C_j$ has misclassified a data example (vector) $\mathbf{x}_i$ and 1 otherwise (if the classifier predicted the class label $f(\mathbf{x}_i)$ correctly).

Applied to an example dataset that was taken from \cite{kuncheva2004combining}, the procedure below illustrates how the classification results may be organized. For instance, assume we have the ground truth labels of the test dataset $\mathbf{y}_{true}$  

 and the following predictions on the test set by 3 classifiers ($\mathbf{y}_{C_1}$, $\mathbf{y}_{C_2}$, and $\mathbf{y}_{C_3}$):

\begin{equation}
\begin{split}
\mathbf{y}_{true} = [0, 0, 0, 0, 0, 0, 0, 0, 0, 0, 0, 0, 0, 0, 0, 0, 0, 0, 0, 0,\\ 
              0, 0, 0, 0, 0, 0, 0, 0, 0, 0, 0, 0, 0, 0, 0, 0, 0, 0, 0, 0, \\
              0, 0, 0, 0, 0, 0, 0, 0, 0, 0, 0, 0, 0, 0, 0, 0, 0, 0, 0, 0, \\
              0, 0, 0, 0, 0, 0, 0, 0, 0, 0, 0, 0, 0, 0, 0, 0, 0, 0, 0, 0,\\
               0, 0, 0, 0, 0, 0, 0, 0, 0, 0, 0, 0, 0, 0, 0, 0, 0, 0, 0, 0];
\end{split}
\end{equation}

\begin{equation}
\begin{split}
\mathbf{y}_{C_1} = [1, 1, 1, 1, 1, 1, 1, 1, 1, 1, 1, 1, 1, 1, 1, 1, 0, 0, 0, 0,\\
0, 0, 0, 0, 0, 0, 0, 0, 0, 0, 0, 0, 0, 0, 0, 0, 0, 0, 0, 0,\\
0, 0, 0, 0, 0, 0, 0, 0, 0, 0, 0, 0, 0, 0, 0, 0, 0, 0, 0, 0,\\
0, 0, 0, 0, 0, 0, 0, 0, 0, 0, 0, 0, 0, 0, 0, 0, 0, 0, 0, 0,\\
0, 0, 0, 0, 0, 0, 0, 0, 0, 0, 0, 0, 0, 0, 0, 0, 0, 0, 0, 0];
\end{split}
\end{equation}

\begin{equation}
\begin{split}
\mathbf{y}_{C_2} = [1, 1, 1, 1, 1, 1, 0, 0, 0, 0, 0, 0, 0, 0, 0, 0, 0, 0, 0, 0,\\
               1, 1, 0, 0, 0, 0, 0, 0, 0, 0, 0, 0, 0, 0, 0, 0, 0, 0, 0, 0,\\
               0, 0, 0, 0, 0, 0, 0, 0, 0, 0, 0, 0, 0, 0, 0, 0, 0, 0, 0, 0,\\
               0, 0, 0, 0, 0, 0, 0, 0, 0, 0, 0, 0, 0, 0, 0, 0, 0, 0, 0, 0,\\
               0, 0, 0, 0, 0, 0, 0, 0, 0, 0, 0, 0, 0, 0, 0, 0, 0, 0, 0, 0];
\end{split}
\end{equation}

\begin{equation}
\begin{split}
\mathbf{y}_{C_3} = [1, 1, 1, 0, 0, 0, 1, 0, 0, 0, 0, 0, 0, 0, 0, 0, 0, 0, 0, 0,\\
               1, 1, 0, 0, 0, 0, 0, 0, 0, 0, 0, 0, 0, 0, 0, 0, 0, 0, 0, 0,\\
               0, 0, 0, 0, 0, 0, 0, 0, 0, 0, 0, 0, 0, 0, 0, 0, 0, 0, 0, 0,\\
               0, 0, 0, 0, 0, 0, 0, 0, 0, 0, 0, 0, 0, 0, 0, 0, 0, 0, 0, 0,\\
               0, 0, 0, 0, 0, 0, 0, 0, 0, 0, 0, 0, 0, 0, 0, 0, 0, 0, 1, 1].
\end{split}
\end{equation}

We can then tabulate the correct (1) and incorrect (0) classifications as shown in Table \ref{tab:cochrans-q}.

\begin{table}[]
\centering
\caption{Table comparing model performances of 3 classifiers used to illustrate the computation of Cochran's $Q$ in Equation \ref{eq:q}}.
\begin{tabular}{lllll}
\hline
         & $C_1$               & $C_2$               & $C_3$               & Occurrences \\ \midrule
         & 1                            & 1                            & 1                            & 80          \\ 
         & 1                            & 1                            & 0                            & 2           \\ 
         & 1                            & 0                            & 1                            & 0           \\ 
         & 1                            & 0                            & 0                            & 2           \\ 
         & 0                            & 1                            & 1                            & 9           \\ 
         & 0                            & 1                            & 0                            & 1           \\ 
         & 0                            & 0                            & 1                            & 3           \\ 
         & 0                            & 0                            & 0                            & 3           \\ \midrule
         & Classification Accuracies:\\\\
          & $84/100 \times 100\% = 84\%$ & $92/100 \times 100\% = 92\%$ & $92/100 \times 100\% = 92\%$ & \\
          \midrule
\end{tabular}
\label{tab:cochrans-q}
\end{table}

By plugging in the respective value into the previous equation, we obtain the following $Q$ value:
\begin{equation}
Q = 2 \times \frac{3 \times (84^2 + 92^2 + 92^2) - 268^2}{3\times 268-(80 \times 9 + 11 \times 4 + 6 \times 1)} \approx 7.5294.
\label{eq:q}
\end{equation}

Now, the $Q$ value (approximating $\chi^2$) corresponds to a $p$-value of approximately 0.023 assuming a $\chi^2$ distribution with $M-1 = 2$ degrees of freedom. Assuming that we chose a significance level of $\alpha=0.05$, we would reject the null hypothesis that all classifiers perform equally well, since $0.023 < \alpha$. (An implementation of Cochran's $Q$ test is provided in MLxtend \cite{raschka2018mlxtend}.\footnote{\url{http://rasbt.github.io/mlxtend/user_guide/evaluate/cochrans_q/}})

In practice, if we successfully rejected the null hypothesis, we could perform multiple post hoc pair-wise tests -- for example, McNemar's test with a Bonferroni correction -- to determine which pairs have different population proportions. Unfortunately, numerous comparisons are usually very tricky in practice. Peter H. Westfall, James F. Troendl, and Gene Pennello wrote a nice article on how to approach such situations where we want to compare multiple models to each other \cite{westfall2010multiple}. 

As Perneger, Thomas V \cite{perneger1998s} writes:

\begin{displayquote}
Type I errors [False Positives] cannot decrease (the whole point of Bonferroni adjustments) without inflating type II errors (the probability of accepting the null hypothesis when the alternative is true) \cite{rothman1990no}. And type II errors [False Negatives] are no less false than type I errors. 
\end{displayquote}

Eventually, once more it comes down to the "no free lunch" -- in this context, let us refer of it as the "no free lunch theorem of statistical tests." However, statistical testing frameworks can be a valuable aid in decision making. So, in practice, if we are honest and rigorous, the process of multiple hypothesis testing with appropriate corrections can be a useful aid in decision making. However, we have to be careful that we do not put too much emphasis on such procedures when it comes to assessing evidence in data.

\subsection{The $F$-test for Comparing Multiple Classifiers}

Almost ironically, Cochran noted that in his work on the $Q$ test \cite{cochran1950comparison}

\begin{displayquote} If the data had been measured variables that appeared normally distributed, instead of a collection of 1's and 0's, the $F$-test would be almost automatically applied as the appropriate method. Without having looked into that matter, I had once or twice suggested to research workers that the $F$-test might serve as an approximation even when the table consists of 1's and 0's	
\end{displayquote}

The method of using the $F$-test for comparing two classifiers in this section is somewhat loosely based on Looney \cite{looney1988statistical}, whereas it shall be noted that Looney recommends an adjusted version called $F^+$ test.

In the context of the $F$-test, our null hypothesis is again that there that there is no difference between the classification accuracies: 

$$
p_i: H_0 = p_1 = p_2 = \cdots = p_L.
$$

Let $\{C_1, \dots , C_M\}$ be a set of classifiers which have all been tested on the same dataset. If the $M$ classifiers do not perform differently, then the F statistic is distributed according to an F distribution with $(M-1$) and $(M-1)\times n$ degrees of freedom, where $n$ is the number of examples in the test set. The calculation of the F statistic consists of several components, which are listed below (adopted from \cite{looney1988statistical}).

We start by defining  $ACC_{avg}$ as the average of the accuracies of the different models

$$ACC_{avg} = \frac{1}{M}\sum_{j=1}^M ACC_j.$$

The sum of squares of the classifiers is then computed as

$$
SSA = n \sum_{j=1}^{M} (G_j)^2 -n \cdot M \cdot ACC_{avg},
$$

where $G_j$ is the proportion of the $n$ examples classified correctly by classifier $j$.

The sum of squares for the objects is calculated as follows:

$$
SSB= \frac{1}{M} \sum_{j=1}^n (M_j)^2 - M\cdot n \cdot ACC_{avg}^2.
$$

Here, $M_j$ is the number of classifiers out of $M$ that correctly classified object $\mathbf{x}_j \in \mathbf{X}_{n}$, where $\mathbf{X}_{n} = \{\mathbf{x}_1, ... \mathbf{x}_{n}\}$ is the test dataset on which the classifiers are tested on.

Finally, we compute the total sum of squares,

$$
SST = M\cdot n \cdot ACC_{avg} (1 - ACC_{avg}),
$$

so that we then can compute the sum of squares for the classification--object interaction:

$$
SSAB = SST - SSA - SSB.
$$

To compute the F statistic, we next compute the mean SSA and mean SSAB values:

$$
MSA = \frac{SSA}{M-1},
$$

and

$$
MSAB = \frac{SSAB}{(M-1) (n-1)}.
$$

From the MSA and MSAB, we can then calculate the $F$-value as

$$
F = \frac{MSA}{MSAB}.
$$

After computing the $F$-value, we can then look up the $p$-value from an F-distribution table for the corresponding degrees of freedom or obtain it computationally from a cumulative F-distribution function. In practice, if we successfully rejected the null hypothesis at a previously chosen significance threshold, we could perform multiple \textit{post hoc} pair-wise tests -- for example, McNemar tests with a Bonferroni correction -- to determine which pairs have different population proportions.

An implementation of this $F$-test is available via MLxtend \cite{raschka2018mlxtend}.\footnote{\url{http://rasbt.github.io/mlxtend/user_guide/evaluate/ftest}}

\subsection{Comparing Algorithms}

The previously described statistical tests focused on model comparisons and thus do not take the variance of the training sets into account, which can be an issue especially if training sets are small and learning algorithms are susceptible to perturbations in the training sets. 

However, if we consider the comparison between sets of models where each set has been fit to different training sets, we conceptually shift from a model compared to an algorithm comparison task. This is often desirable, though. For instance, assume we develop a new learning algorithm or want to decide which learning algorithm is best to ship with our new software (a trivial example would be an email program with a learning algorithm that learns how to filter spam based on the user's decisions). In this case, we are want to find out how different algorithms perform on datasets from a similar problem domain. 

One of the commonly used techniques for algorithm comparison is Thomas Dietterich's 5x2-Fold Cross-Validation method (5x2cv for short) that was introduced in his paper "Approximate statistical tests for comparing supervised classification learning algorithms" \cite{dietterich1998approximate}. It is a nice paper that discusses all the different testing scenarios (the different circumstances and applications for model evaluation, model selection, and algorithm selection) in the context of statistical tests. The conclusions that can be drawn from empirical comparison on simulated datasets are summarized below.

\begin{enumerate}
\item  McNemar's test:
   \begin{itemize}
   \item low false positive rate
   \item fast, only needs to be executed once
   \end{itemize}
\item  The difference in proportions test:
	\begin{itemize}
     \item high false positive rate (here, incorrectly detected a difference when there is none)
     \item cheap to compute, though
   \end{itemize}
\item  Resampled paired $t$-test:
   \begin{itemize}
   \item high false positive rate
   \item computationally very expensive
   \end{itemize}
\item  $k$-fold cross-validated $t$-test:
   \begin{itemize}
   \item somewhat elated false positive rate
   \item requires refitting to training sets; $k$ times more computations than McNemar's test
   \end{itemize}
\item  5x2cv paired $t$-test
   \begin{itemize}
   \item low false positive rate (similar to McNemar's test)
   \item slightly more powerful than McNemar's test; recommended if computational efficiency (runtime) is not an issue (10 times more computations than McNemar's test)
   \end{itemize}
\end{enumerate}

The bottom line is that McNemar's test is a good choice if the datasets are relatively large and/or the model fitting can only be conducted once. If the repeated fitting of the models is possible, the 5x2cv test is a good choice as it also considers the effect of varying or resampled training sets on the model fitting. 

For completeness, the next section will summarize the mechanics of the tests we have not covered thus far.

\subsection{Resampled Paired $t$-Test}

Resampled paired $t$-test procedure (also called k-hold-out paired $t$-test) is a popular method for comparing the performance of two models (classifiers or regressors); however, this method has many drawbacks and is not recommended to be used in practice as Dietterich \cite{dietterich1998approximate} noted.

To explain how this method works, let us consider two classifiers $C_1$ and $C_2$. Further, we have a labeled dataset $S$. In the common hold-out method, we typically split the dataset into 2 parts: a training and a test set. In the resampled paired $t$-test procedure, we repeat this splitting procedure (with typically 2/3 training data and 1/3 test data) $k$ times (usually 30 or more). In each iteration, we fit both $C_1$ and $C_2$ on the same training set and evaluate these on the same test set. Then, we compute the difference in performance between $C_1$ and $C_2$ in each iteration so that we obtain $k$ difference measures. Now, by making the assumption that these $k$ differences were independently drawn and follow an approximately normal distribution, we can compute the following $t$ statistic with $k-1$ degrees of freedom according to Student's $t$-test, under the null hypothesis that the models $C_1$ and $C_2$ have equal performance:
$$
t = \frac{ACC_{avg} \sqrt{k}}{\sqrt{\sum_{i=1}^{k}(ACC_{i} - ACC_{avg})^2 / (k-1)}}.
$$
Here, $ACC_i$ computes the difference between the model accuracies in the $i$th iteration $ACC_i = ACC_{i, C_1} - ACC_{i, C_2}$ and $ACC_{avg}$ represents the average difference between the classifier performances $ACC_{avg} = \frac{1}{k} \sum_{i=1}^k ACC_i $.

Once we computed the $t$ statistic, we can calculate the $p$-value and compare it to our chosen significance level, for example, $\alpha=0.05$. If the $p$-value is smaller than $\alpha$, we reject the null hypothesis and accept that there is a significant difference between the two models.

The problem with this method, and the reason why it is not recommended to be used in practice, is that it violates the assumptions of Student's $t$-test, as the differences of the model performances are not normally distributed because the accuracies are not independent (since we compute them on the same test set). Also, the differences between the accuracies themselves are also not independent since the test sets overlap upon resampling. Hence, in practice, it is not recommended to use this test in practice. However, for comparison studies, this test is implemented in MLxtend\cite{raschka2018mlxtend}.\footnote{\url{http://rasbt.github.io/mlxtend/user_guide/evaluate/paired_ttest_resampled/}}.

\subsection{$k$-fold Cross-validated Paired $t$-Test}

Similar to the resampled paired $t$-test, the $k$-fold cross-validated paired $t$-test is a statistical testing technique that is very common in (older) literature.  While it addresses some of the drawbacks of the resampled paired $t$-test procedure, this method has still the problem that the training sets overlap and is hence also not recommended to be used in practice \cite{dietterich1998approximate}. 

Again, for completeness, the method is outlined below. The procedure is basically equivalent to that of the resampled paired $t$-test procedure except that we use $k$-fold cross validation instead of simple resampling, such that if we compute the $t$ value,

$$
t = \frac{ACC_{avg} \sqrt{k}}{\sqrt{\sum_{i=1}^{k}(ACC_{i} - ACC_{avg})^2 / (k-1)}},
$$

$k$ is equal to the number of cross-validation rounds. Again, for comparison studies, I made this this testing procedure available through MLxtend \cite{raschka2018mlxtend}.\footnote{\url{http://rasbt.github.io/mlxtend/user_guide/evaluate/paired_ttest_kfold_cv/}}

\subsection{Dietterich's 5x2-Fold Cross-Validated Paired $t$-Test}

The 5x2cv paired $t$-test is a procedure for comparing the performance of two models (classifiers or regressors) that was proposed by Dietterich (Dietterich, 1998) to address shortcomings in other methods such as the resampled paired $t$-test and the $k$-fold cross-validated paired $t$-test, which were outlined in the previous two sections.

While the overall approach is similar to the previously described $t$-test variants, in the 5x2cv paired $t$-test, we repeat the splitting (50\% training and 50\% test data) five times. 

In each of the 5 iterations, we fit two classifiers $C_1$ and $C_2$ to the training split and evaluate their performance on the test split. Then, we rotate the training and test sets (the training set becomes the test set and vice versa) compute the performance again, which results in two performance difference measures:
$$
ACC_A = ACC_{A, C_1} - ACC_{A, C_2},
$$

and

$$
ACC_B = ACC_{B, C_1} - ACC_{B, C_2}.
$$

Then, we estimate the estimate mean and variance of the differences:

$$
ACC_{avg} = (ACC_A + ACC_B) / 2,
$$

and

$$
s^2 = (ACC_A - ACC_{avg})^2 + (ACC_B - ACC_{avg}^2)^2.
$$

The variance of the difference is computed for the 5 iterations and then used to compute the $t$ statistic as follows:
$$
t = \frac{ACC_{A, 1}}{\sqrt{(1/5) \sum_{i=1}^{5}s_i^2}},
$$

where $ACC_{A, 1}$ is the $ACC_{A}$ obtained from the first iteration.

The $t$ statistic approximately follows as $t$ distribution with 5 degrees of freedom, under the null hypothesis that the models $C_1$ and $C_2$ have equal performance. Using the $t$ statistic, the $p$-value can then be computed and compared with a previously chosen significance level, for example, $\alpha=0.05$. If the $p$-value is smaller than $\alpha$, we reject the null hypothesis and accept that there is a significant difference in the two models. The 5x2cv paired $t$-test is available from MLxtend \cite{raschka2018mlxtend}.\footnote{\url{http://rasbt.github.io/mlxtend/user_guide/evaluate/paired_ttest_5x2cv/}}

\subsection{Alpaydin's Combined 5x2cv $F$-test}

The 5x2cv combined $F$-test is a procedure for comparing the performance of models (classifiers or regressors)
that was proposed by Alpaydin \cite{alpaydin1999combined} as a more robust alternative to Dietterich's 5x2cv paired $t$-test procedure outlined in the previous section. 

To explain how this mechanics of this method, let us consider two classifiers 1 and 2 and re-use the notation from the previous section. The F statistic is then computed as follows:
$$
f = \frac{\sum_{i=1}^{5} \sum_{j=1}^2 (ACC_{i,j})^2}{2 \sum_{i=1}^5 s_i^2},
$$

which is approximately $F$ distributed with 10 and 5 degrees of freedom.  The combined 5x2cv $F$-test is available from MLxtend \cite{raschka2018mlxtend}.\footnote{\url{http://rasbt.github.io/mlxtend/user_guide/evaluate/combined_ftest_5x2cv/}}.

\subsection{Effect size}

While (unfortunately rarely done in practice), we may also want to consider effect sizes since large samples elevate $p$-values and can make \textit{everything} seem statistically significant. In other words, "theoretical significance" does not imply "practical significance." As effect size is a more of an objective topic that depends on the problem/task/question at hand, a detailed discussion is obviously out of the scope of this article.

\subsection{Nested Cross-Validation}

In practical applications, we usually never have the luxury of having a large (or, ideally infinitely) sized test set, which would provide us with an unbiased estimate of the true generalization error of a model. Hence, we are always on a quest of finding "better" workaround for dealing with size-limited datasets: Reserving too much data for training results in unreliable estimates of the generalization performance, and setting aside too much data for testing results in too little data for training, which hurts model performance.

Almost always, we also do not know the ideal settings of the learning algorithm for a given problem or problem domain. Hence, we need to use an available training set for hyperparameter tuning and model selection. We established earlier that we could use $k$-fold cross-validation as a method for these tasks. However, if we select the "best hyperparameter settings" based on the average $k$-fold performance or the *same* test set, we introduce a bias into the procedure, and our model performance estimates will not be unbiased anymore. Mainly, we can think of model selection as another \textit{training} procedure, and hence, we would need a decently-sized, independent test set that we have not seen before to get an unbiased estimate of the models' performance. Often, this is not affordable.

In recent years, a technique called \textit{nested cross-validation} has emerged as one of the popular or somewhat recommended methods for comparing machine learning algorithms; it was likely first described by Iizuka \cite{iizuka2003oligonucleotide} and Varma and Simon \cite{varma2006bias} when working with small datasets. The nested cross-validation procedure offers a workaround for small-dataset situations that shows a low bias in practice where reserving data for independent test sets is not feasible. 

Varma and Simon found that the nested cross-validation approach can reduce the bias, compared to regular $k$-fold cross-validation when used for both hyperparameter tuning and evaluation, can be considerably be reduced. As the researchers state, "A nested CV procedure provides an almost unbiased estimate of the true error" \cite{varma2006bias}.

\begin{figure}[htb!]
 \centering
    \includegraphics[width=\linewidth]{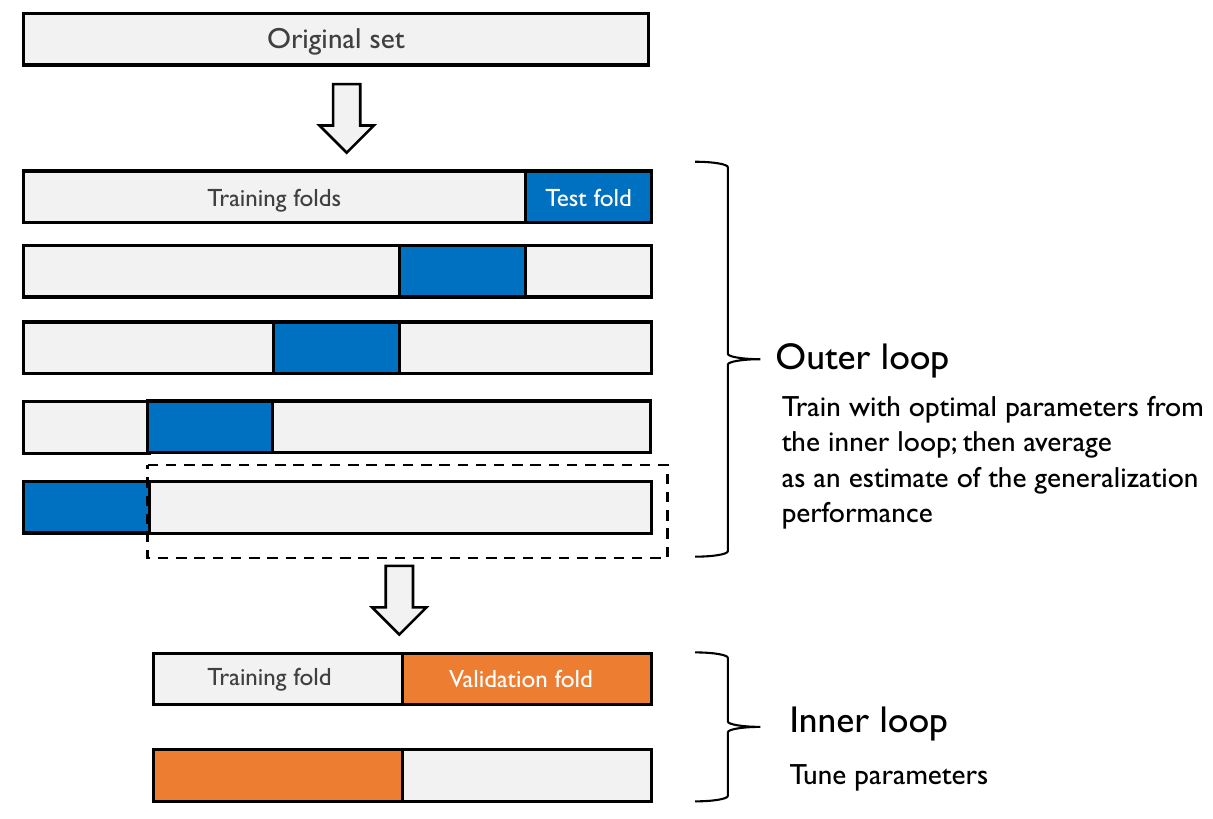}
    \caption{Illustration of nested cross-validation.}
    \label{fig21:nested-cv}
\end{figure}

The method of nested cross-validation is relatively straight-forward as it merely is a nesting of two $k$-fold cross-validation loops: the inner loop is responsible for the model selection, and the outer loop is responsible for estimating the generalization accuracy, as shown in Figure \ref{fig21:nested-cv}.

Note that in this particular case, this is a 5x2 setup (5-fold cross-validation in the outer loop, and 2-fold cross-validation in the inner loop). However, this is not the same as Dietterich's 5x2cv method, which is an often confused scenario such that I want to highlight it here. 

Code for using nested cross-validation with scikit-learn \cite{pedregosa2011scikit} can be found at \url{https://github.com/rasbt/model-eval-article-supplementary/blob/master/code/nested_cv_code.ipynb}.

\subsection{Conclusions}

\begin{figure}[htb!]
 \centering
    \includegraphics[width=\linewidth]{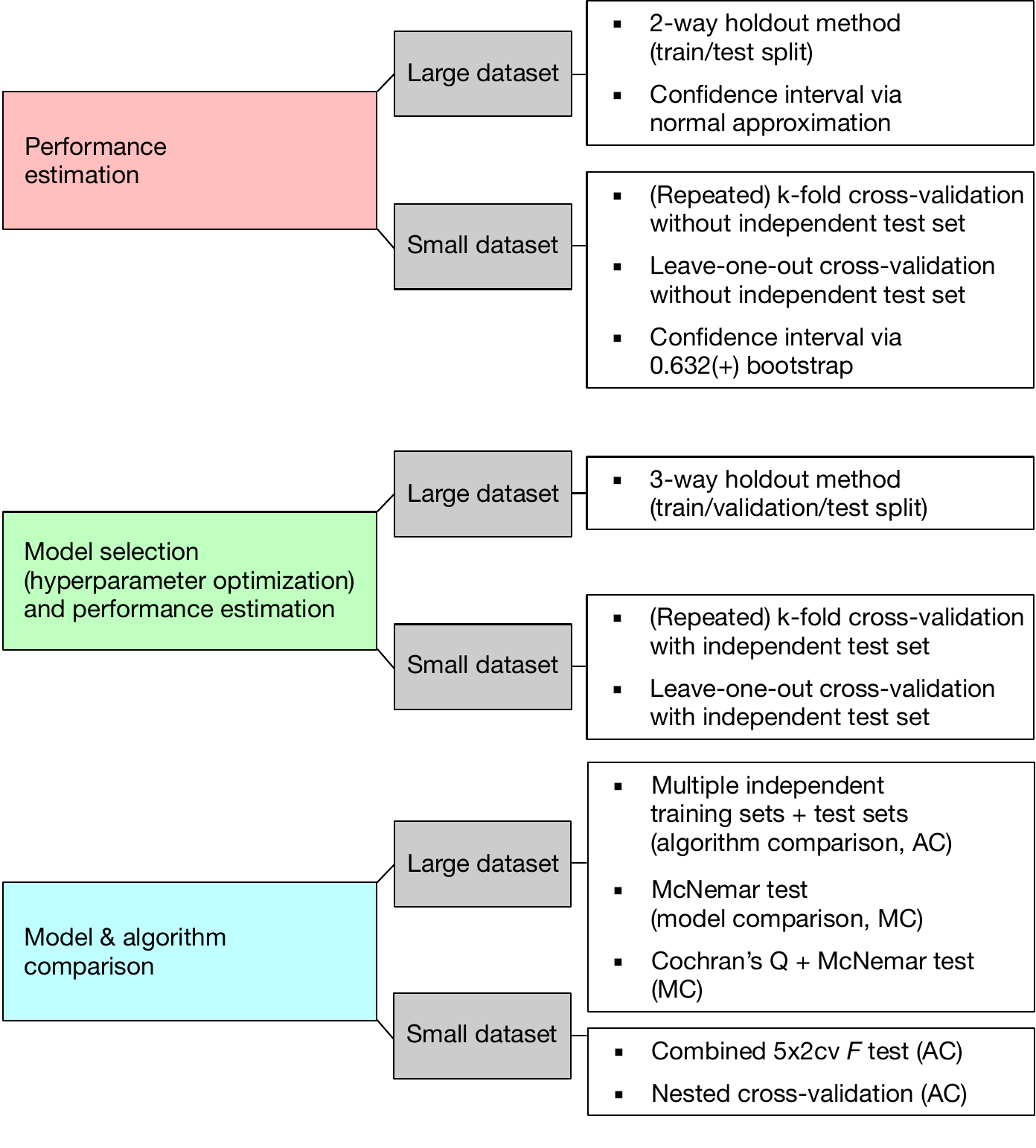}
    \caption{A recommended subset of techniques to be used to address different aspects of model evaluation in the context of small and large datasets. The abbreviation "MC" stands for "Model Comparison," and "AC" stands for "Algorithm Comparison," to distinguish these two tasks.}
    \label{fig22:conclusions}
\end{figure}

Since "a picture is worth a thousand words," I want to conclude this series on model evaluation, model selection, and algorithm selection with a diagram (Figure \ref{fig22:conclusions}) that summarizes my personal recommendations based on the concepts and literature that was reviewed.

It should be stressed that parametric tests for comparing model performances usually violate one or more independent assumptions (the models are not independent because the same training set was used, and the estimated generalization performances are not independent because the same test set was used.). In an ideal world, we would have access to the data generating distribution or at least an almost infinite pool of new data. However, in most practical applications, the size of the dataset is limited; hence, we can use one of the statistical tests discussed in this article as a heuristic to aid our decision making.

Note that the recommendations I listed in the figure above are suggestions and depend on the problem at hand. For instance, large test datasets (where "large" is relative but might refer to thousands or millions of data records), can provide reliable estimates of the generalization performance, whereas using a single training and test set when only a few data records are available can be problematic for several reasons discussed throughout Section \ref{sec:boostrapping-and-uncertainties} and Section \ref{sec:cross-val}. If the dataset is very small, it might not be feasible to set aside data for testing, and in such cases, we can use $k$-fold cross-validation with a large $k$ or Leave-one-out cross-validation as a workaround for evaluating the generalization performance. However, using these procedures, we have to bear in mind that we then do not compare between models but different algorithms that produce different models on the training folds. Nonetheless, the average performance over the different test folds can serve as an estimate for the generalization performance (Section \ref{sec:cross-val}) discussed the various implications for the bias and the variance of this estimate as a function of the number of folds).

For model comparisons, we usually do not have multiple independent test sets to evaluate the models on, so we can again resort to cross-validation procedures such as $k$-fold cross-validation, the 5x2cv method, or nested cross-validation.  As Gael Varoquaux \cite{varoquaux2017cross} writes:

\begin{displayquote} Cross-validation is not a silver bullet. However, it is the best tool available, because it is the only non-parametric method to test for model generalization. \end{displayquote}

\subsection{Acknowledgments}

I would like to thank Simon Opstrup Drue for carefully reading the manuscript and providing useful suggestions.

\bibliographystyle{apalike}

\begin{thebibliography}{}

\end{thebibliography}


\begin{thebibliography}{}

\bibitem[Alpaydin, 1999]{alpaydin1999combined}
Alpaydin, E. (1999).
\newblock Combined 5x2cv {F} test for comparing supervised classification
  learning algorithms.
\newblock {\em Neural Computation}, 11(8):1885--1892.

\bibitem[Bengio and Grandvalet, 2004]{bengio2004no}
Bengio, Y. and Grandvalet, Y. (2004).
\newblock No unbiased estimator of the variance of k-fold cross-validation.
\newblock {\em Journal of Machine Learning Research}, 5(Sep):1089--1105.

\bibitem[Bonferroni, 1936]{bonferroni1936teoria}
Bonferroni, C. (1936).
\newblock Teoria statistica delle classi e calcolo delle probabilita.
\newblock {\em Pubblicazioni del R Istituto Superiore di Scienze Economiche e
  Commericiali di Firenze}, 8:3--62.

\bibitem[Breiman et~al., 1984]{breiman1984classification}
Breiman, L., Friedman, J., Stone, C.~J., and Olshen, R.~A. (1984).
\newblock {\em Classification and regression trees}.
\newblock CRC press.

\bibitem[Cochran, 1950]{cochran1950comparison}
Cochran, W.~G. (1950).
\newblock The comparison of percentages in matched samples.
\newblock {\em Biometrika}, 37(3/4):256--266.

\bibitem[Dietterich, 1998]{dietterich1998approximate}
Dietterich, T.~G. (1998).
\newblock Approximate statistical tests for comparing supervised classification
  learning algorithms.
\newblock {\em Neural computation}, 10(7):1895--1923.

\bibitem[Dunn, 1961]{dunn1961multiple}
Dunn, O.~J. (1961).
\newblock Multiple comparisons among means.
\newblock {\em Journal of the American statistical association},
  56(293):52--64.

\bibitem[Edwards, 1948]{edwards1948note}
Edwards, A.~L. (1948).
\newblock Note on the “correction for continuity” in testing the
  significance of the difference between correlated proportions.
\newblock {\em Psychometrika}, 13(3):185--187.

\bibitem[Efron, 1981]{efron1981nonparametric}
Efron, B. (1981).
\newblock Nonparametric standard errors and confidence intervals.
\newblock {\em Canadian Journal of Statistics}, 9(2):139--158.

\bibitem[Efron, 1983]{efron1983estimating}
Efron, B. (1983).
\newblock Estimating the error rate of a prediction rule: improvement on
  cross-validation.
\newblock {\em Journal of the American Statistical Association},
  78(382):316--331.

\bibitem[Efron, 1992]{efron1992bootstrap}
Efron, B. (1992).
\newblock Bootstrap methods: another look at the {Jackknife}.
\newblock In {\em Breakthroughs in Statistics}, pages 569--593. Springer.

\bibitem[Efron and Tibshirani, 1997]{efron1997improvements}
Efron, B. and Tibshirani, R. (1997).
\newblock Improvements on cross-validation: the .632+ bootstrap method.
\newblock {\em Journal of the American Statistical Association},
  92(438):548--560.

\bibitem[Efron and Tibshirani, 1994]{efron1994introduction}
Efron, B. and Tibshirani, R.~J. (1994).
\newblock {\em An Introduction to the Bootstrap}.
\newblock CRC press.

\bibitem[Fleiss et~al., 2013]{fleiss2013statistical}
Fleiss, J.~L., Levin, B., and Paik, M.~C. (2013).
\newblock {\em Statistical Methods for Rates and Proportions}.
\newblock John Wiley \& Sons.

\bibitem[Hastie et~al., 2009]{hastie2009}
Hastie, T., Tibshirani, R., and Friedman, J.~H. (2009).
\newblock {\em In The Elements of Statistical Learning: Data Mining, Inference,
  and Prediction}.
\newblock Springer, New York.

\bibitem[Hawkins et~al., 2003]{hawkins2003assessing}
Hawkins, D.~M., Basak, S.~C., and Mills, D. (2003).
\newblock Assessing model fit by cross-validation.
\newblock {\em Journal of Chemical Information and Computer Sciences},
  43(2):579--586.

\bibitem[Iizuka et~al., 2003]{iizuka2003oligonucleotide}
Iizuka, N., Oka, M., Yamada-Okabe, H., Nishida, M., Maeda, Y., Mori, N., Takao,
  T., Tamesa, T., Tangoku, A., Tabuchi, H., et~al. (2003).
\newblock Oligonucleotide microarray for prediction of early intrahepatic
  recurrence of hepatocellular carcinoma after curative resection.
\newblock {\em The lancet}, 361(9361):923--929.

\bibitem[James et~al., 2013]{gareth2013stat}
James, G., Witten, D., Hastie, T., and Tibshirani, R. (2013).
\newblock {\em In An Introduction to Statistical Learning: With Applications in
  R}.
\newblock Springer, New York.

\bibitem[Kim, 2009]{kim2009estimating}
Kim, J.-H. (2009).
\newblock Estimating classification error rate: Repeated cross-validation,
  repeated hold-out and bootstrap.
\newblock {\em Computational Statistics \& Data Analysis}, 53(11):3735--3745.

\bibitem[Kohavi, 1995]{kohavi1995}
Kohavi, R. (1995).
\newblock {A study of cross-validation and bootstrap for accuracy estimation
  and model selection}.
\newblock {\em International Joint Conference on Artificial Intelligence},
  14(12):1137--1143.

\bibitem[Kuncheva, 2004]{kuncheva2004combining}
Kuncheva, L.~I. (2004).
\newblock {\em Combining Pattern Classifiers: Methods and Algorithms}.
\newblock John Wiley \& Sons.

\bibitem[Looney, 1988]{looney1988statistical}
Looney, S.~W. (1988).
\newblock A statistical technique for comparing the accuracies of several
  classifiers.
\newblock {\em Pattern Recognition Letters}, 8(1):5--9.

\bibitem[McNemar, 1947]{mcnemar1947note}
McNemar, Q. (1947).
\newblock Note on the sampling error of the difference between correlated
  proportions or percentages.
\newblock {\em Psychometrika}, 12(2):153--157.

\bibitem[Molinaro et~al., 2005]{molinaro2005prediction}
Molinaro, A.~M., Simon, R., and Pfeiffer, R.~M. (2005).
\newblock Prediction error estimation: a comparison of resampling methods.
\newblock {\em Bioinformatics}, 21(15):3301--3307.

\bibitem[Pedregosa et~al., 2011]{pedregosa2011scikit}
Pedregosa, F., Varoquaux, G., Gramfort, A., Michel, V., Thirion, B., Grisel,
  O., Blondel, M., Prettenhofer, P., Weiss, R., Dubourg, V., et~al. (2011).
\newblock Scikit-learn: Machine learning in python.
\newblock {\em Journal of Machine Learning Research}, 12(Oct):2825--2830.

\bibitem[Perneger, 1998]{perneger1998s}
Perneger, T.~V. (1998).
\newblock What's wrong with bonferroni adjustments.
\newblock {\em Bmj}, 316(7139):1236--1238.

\bibitem[Raschka, 2018]{raschka2018mlxtend}
Raschka, S. (2018).
\newblock Mlxtend: Providing machine learning and data science utilities and
  extensions to python’s scientific computing stack.
\newblock {\em The Journal of Open Source Software}, 3(24).

\bibitem[Refaeilzadeh et~al., 2007]{refaeilzadeh2007comparison}
Refaeilzadeh, P., Tang, L., and Liu, H. (2007).
\newblock {On comparison of feature selection algorithms}.
\newblock In {\em Proceedings of AAAI Workshop on Evaluation Methods for
  Machine Learning II}, pages 34--39.

\bibitem[Rothman, 1990]{rothman1990no}
Rothman, K.~J. (1990).
\newblock No adjustments are needed for multiple comparisons.
\newblock {\em Epidemiology}, pages 43--46.

\bibitem[Tan et~al., 2005]{tan2005datamining}
Tan, P.-N., Steinbach, M., and Kumar, V. (2005).
\newblock {\em In Introduction to Data Mining}.
\newblock Pearson Addison Wesley, Boston.

\bibitem[Varma and Simon, 2006]{varma2006bias}
Varma, S. and Simon, R. (2006).
\newblock Bias in error estimation when using cross-validation for model
  selection.
\newblock {\em BMC bioinformatics}, 7(1):91.

\bibitem[Varoquaux, 2017]{varoquaux2017cross}
Varoquaux, G. (2017).
\newblock Cross-validation failure: small sample sizes lead to large error
  bars.
\newblock {\em Neuroimage}.

\bibitem[Westfall et~al., 2010]{westfall2010multiple}
Westfall, P.~H., Troendle, J.~F., and Pennello, G. (2010).
\newblock Multiple {McNemar} tests.
\newblock {\em Biometrics}, 66(4):1185--1191.

\end{thebibliography}

\end{document}